\documentclass{article}
\usepackage[T1]{fontenc}
\usepackage{iclr2026_conference,times}

\usepackage{amsmath,amsfonts,bm}

\def\eqref#1{equation~\ref{#1}}

\def\1{\bm{1}}

\DeclareMathAlphabet{\mathsfit}{\encodingdefault}{\sfdefault}{m}{sl}
\SetMathAlphabet{\mathsfit}{bold}{\encodingdefault}{\sfdefault}{bx}{n}

\usepackage{latexsym}
\usepackage{amsmath}
\usepackage{pifont}
\usepackage{amsfonts}
\usepackage{colortbl}
\usepackage{tcolorbox}
\usepackage{amssymb}
\usepackage{graphicx}
\usepackage{booktabs}
\usepackage{array}
\usepackage{multirow}
\usepackage{tabularx}
\usepackage{listings}
\usepackage{courier}
\usepackage[table]{xcolor}
\usepackage{hyperref}  
\usepackage{pifont}  
\usepackage{subcaption}
\usepackage{listings}
\newcommand{\cmark}{\textcolor{green!60!black}{\ding{51}}} 
\newcommand{\xmark}{\textcolor{red}{\ding{55}}}

\lstdefinestyle{prompt}{
  basicstyle=\linespread{1.05}\ttfamily\small,
  columns=fullflexible,
  breaklines=true,
  breakatwhitespace=false,
  frame=single,
  rulecolor=\color{black!20},
  frameround=tttt,
  tabsize=2,
  showstringspaces=false,
  upquote=true,
  numbers=left,
  numberstyle=\tiny\color{black!50},
  numbersep=8pt
}
\title{Charts Are Not Images: \\On the Challenges of Scientific Chart Editing}

\author{%
\parbox{\textwidth}{
\textbf{Shawn Li$^1$, Ryan Rossi$^2$, Sungchul Kim$^2$, Sunav Choudhary$^2$, Franck Dernoncourt$^2$}\\
\textbf{Puneet Mathur$^2$, Zhengzhong Tu$^3$, Yue Zhao$^1$}
}\\[4pt]
$^1$University of Southern California, $^2$Adobe Research, $^3$Texas A\&M University\\
\texttt{(li.li02, yue.z)@usc.edu}\\
\texttt{(ryrossi, sukim, schoudha, dernonco, puneetm)@adobe.com}\\
\texttt{tzz@tamu.edu}\\
}

\iclrfinalcopy 

\begin{document}
\maketitle

\begin{abstract}
Generative models, such as diffusion and autoregressive approaches, have demonstrated impressive capabilities in editing natural images. However, applying these tools to scientific charts rests on a flawed assumption: a chart is not merely an arrangement of pixels but a visual representation of structured data governed by a graphical grammar. Consequently, chart editing is not a pixel-manipulation task but a structured transformation problem.
To address this fundamental mismatch, we introduce \textit{FigEdit}, a large-scale benchmark for scientific figure editing comprising over 30,000 samples. Grounded in real-world data, our benchmark is distinguished by its diversity, covering 10 distinct chart types and a rich vocabulary of complex editing instructions. The benchmark is organized into five distinct and progressively challenging tasks: single edits, multi edits, conversational edits, visual-guidance-based edits, and style transfer.
Our evaluation of a range of state-of-the-art models on this benchmark reveals their poor performance on scientific figures, as they consistently fail to handle the underlying structured transformations required for valid edits. Furthermore, our analysis indicates that traditional evaluation metrics (e.g., SSIM, PSNR) have limitations in capturing the semantic correctness of chart edits. Our benchmark demonstrates the profound limitations of pixel-level manipulation and provides a robust foundation for developing and evaluating future structure-aware models. By releasing \textit{FigEdit} (\url{https://github.com/adobe-research/figure-editing}), we aim to enable systematic progress in structure-aware figure editing, provide a common ground for fair comparison, and encourage future research on models that understand both the visual and semantic layers of scientific charts.
\end{abstract}

\section{Introduction}

Vision-language models (VLMs) have advanced rapidly, showing strong results in recognition, captioning, and instruction-following image editing \citep{radford2021learning,schuhmann2022laion5bopenlargescaledataset,rombach2022highresolutionimagesynthesislatent,brooks2023instructpix2pixlearningfollowimage,zhang2023addingconditionalcontroltexttoimage,team2023gemini,openai2024gpt4o,chen2024internvl,wang2024qwen2,lu2024deepseek,liu2024llavanext,li2024llava,yao2024minicpm,xu2024llava}. Beyond natural images, chart editing focuses on the precise modification of charts and graphs from natural-language instructions, which is central to scientific communication and data analysis \cite{litomm,li2025personalizedconversationalbenchmarksimulating}. Typical workflows include updating figures when upstream tables change, adapting layouts for publication, aligning styles across related plots, and converting encodings to highlight specific trends. In collaborative environments, edits often arrive as multi-turn requests with references to earlier messages, related figures, or localized visual cues. Such use cases require outputs that remain faithful to underlying data, consistent with visualization rules, and auditable for provenance \citep{belouadi2024automatikztextguidedsynthesisscientific}. At the same time, instruction-tuning and dialogue-centric editing continue to expand the ability of modern systems to follow multi-turn control \citep{li2024enhanced,huang2024dialoggen,ma2025dialogdraw,wei2024balancing,hahn2024proactive,deng2025proactive,zhang2025survey,li2024panoptic,li2025dpu,li2025treble,liu2025continual,liu2025principled,li2025secure}.

Despite these advances, figure editing differs fundamentally from natural image manipulation. A chart is the rendering of structured data through a graphical grammar, and valid edits are \emph{structured transformations} on marks, scales, encodings, and legends rather than pixel changes. Instructions such as “add a bar for category \emph{X} with value 42” require coherent updates to data schema and visual mappings, yet current models often treat them as visual rearrangements, producing outputs that appear plausible but violate semantics. This exposes a persistent problem–method mismatch: instruction-following editors and multi-turn generation systems \citep{brooks2023instructpix2pixlearningfollowimage,zhang2023addingconditionalcontroltexttoimage,NEURIPS2023_f8ad010c,Wang_2024} are optimized for perceptual alignment under open-ended goals, whereas figure editing is constrained by data fidelity and visualization rules. Models trained on web-scale natural images \citep{schuhmann2022laion5bopenlargescaledataset,radford2021learning} lack inductive bias to preserve value–encoding consistency, axis coherence, and legend integrity. While dialog-driven clarification \citep{andukuri2024star,chen2024learning,zelikman2024star} or OCR augmentation \citep{10030860,rodriguez2023ocr} can mitigate ambiguity locally, they do not guarantee structure-preserving edits, leaving the core mismatch unresolved.

\paragraph{Current approaches and benchmarks.}
On the approach side, diffusion editors and multimodal LLMs have been extended to multi-turn control and retrieval-augmented interaction \citep{li2024enhanced,huang2024dialoggen,ma2025dialogdraw,wei2024balancing,wang2025twin,hahn2024proactive,deng2025proactive,liu2024you,taneja2025mudoc,zhao2025chatsearch}. Yet, these systems rarely operate on executable specifications or enforce semantic constraints, which makes them unsuitable for structured figure editing.
On the benchmark side, prior chart-related datasets have mainly targeted captioning, QA, table extraction, or chart-to-code generation \citep{hsu2021scicap,kantharaj2022chart,masry2023unichart,han2023chartllama,zhang2024tinychart,qin2024metaood,xia2024chartx,shi2024chartmimic,Masry2024ChartInstructIT,zhang2024scimagegoodmultimodallarge}. As shown in Tab.~\ref{tab:benchmark_comparison}, these resources leave several gaps. Some lack real underlying data altogether (e.g., \citealt{xia2024chartx,zhang2024gpt}), reducing their grounding in authentic visualization workflows. Coverage of edit categories is also narrow: data-level updates, layout transformations, and style changes are often missing. Interactive scenarios such as visual guidance or style transfer are almost entirely absent, despite being common in real practice. Even the recent ChartEdit benchmark \citep{zhao2025chartedit}, while closer to editing, only partially spans instruction types and lacks paired figure outputs for direct comparison. Overall, existing benchmarks fall short of representing the breadth of figure editing and still depend heavily on pixel-level similarity metrics, which do not reflect semantic correctness. This highlights the need for a task-structured, semantics-aware, and scale-ready benchmark dedicated to figure editing.

\begin{figure}[t]
\centering
\includegraphics[width=\textwidth]{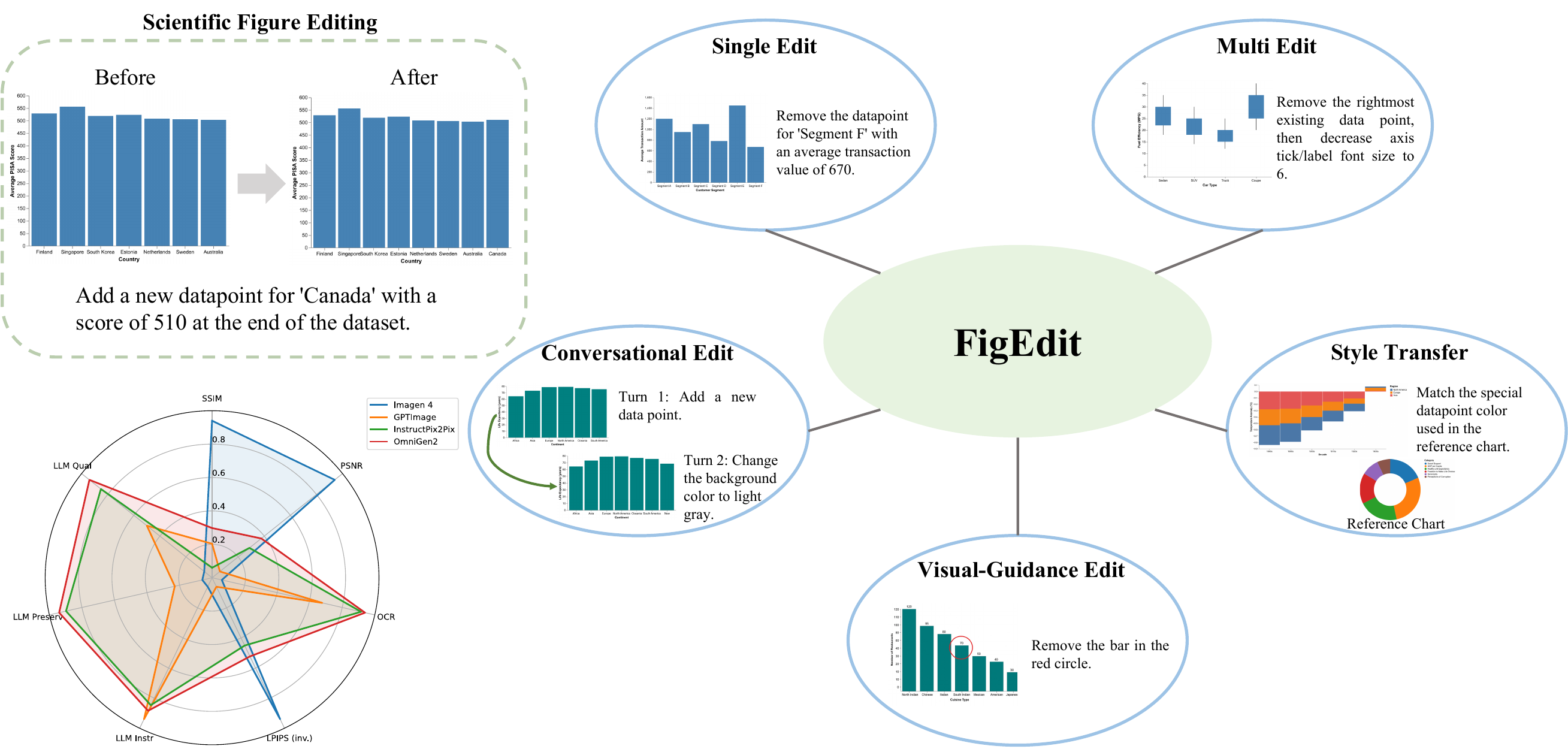} 
\vspace{-0.1in}
   \caption{
    FigEdit benchmark. 
    Top-left: an example figure illustrating the basic task. 
    Bottom-left: a radar chart comparing model performance on single edit task, highlighting the benchmark’s ability to reveal differences in editing capabilities.
    Right: taxonomy of the benchmark covering five tasks (single edit, multi edit, conversational edit, visual guidance, and style transfer).}
\label{fig:face}
\vspace{-0.5cm}
\end{figure}

\paragraph{Our benchmark.}
We introduce \textit{FigEdit}, a large-scale benchmark for scientific chart editing with over 30,000 instances collected from realistic sources (Fig.~\ref{fig:face}). It spans 10 chart types and a diverse set of instructions, as summarized in Tab.~\ref{tab:chart_type_stats}, and is organized into five evaluation settings: single edits, multi edits, conversational edits, visual-guided edits, and style transfer edits. The benchmark also covers a wide range of operation categories, including data-centric edits, layout adjustments, style modifications, and text updates, detailed in Tab.~\ref{tab:ops_summary_all}. Unlike prior benchmarks that lack real data or paired chart outputs, \textit{FigEdit} grounds edits in authentic charts and provides both charts and specification references. To address the absence of interactive scenarios, it includes conversational editing for multi-turn consistency, visual-guided editing with localized cues, and style transfer for cross-chart alignment. Finally, beyond SSIM and PSNR, \textit{FigEdit} introduces semantics-aware evaluation that verifies transformations at the level of data and encodings, with executable targets or programmatic specifications where possible \citep{li2024mmcode,zhang2024humaneval,zheng2023codegeex,wei2024magicoder,guo2024deepseek,shi2024chartmimic}. These design choices directly address the limitations of existing benchmarks and shift evaluation from pixel similarity toward semantic correctness in structured editing.

Our contributions are summarized as follows:
\begin{itemize}
    \item \textbf{Problem formalization}: We define chart editing as a \emph{structured transformation} task governed by a graphical grammar, clarifying required invariants such as data–encoding alignment, axis coherence, and legend integrity.
    \item \textbf{Task-structured benchmark}: We present \textit{FigEdit}, a benchmark with 30K+ instances and 10 chart types, spanning single, multi, conversational, visual-guided, and style transfer with a diverse instruction set.
    \item \textbf{Comprehensive study}: We systematically evaluate state-of-the-art editors and VLMs, showing that strong scores on pixel metrics do not imply correct structured edits, and analyze frequent failure modes.
\end{itemize}

\section{Related Work}

\noindent \textbf{Text-to-Image Generation.}  
Diffusion models have advanced text-conditioned image generation, producing high-fidelity results \citep{ramesh2022hierarchicaltextconditionalimagegeneration,rombach2022highresolutionimagesynthesislatent}. Methods such as ControlNet add controllability via spatial priors \citep{zhang2023addingconditionalcontroltexttoimage}, but these works mainly target natural images. Scientific figures remain underexplored, where symbolic precision and textual fidelity are critical \citep{zhang2024scimagegoodmultimodallarge,rodriguez2023ocr,belouadi2024automatikztextguidedsynthesisscientific}.  

\noindent \textbf{Image Editing.}  
Instruction-driven editing has progressed rapidly with diffusion models, surpassing earlier GAN- or encoder-based approaches in balancing realism and alignment \citep{10884879}. Representative systems include LEDITS++ \citep{Brack_2024_CVPR}, Emu Edit \citep{sheynin2024emu}, and SmartEdit \citep{huang2024smartedit}. Interactive and compositional methods such as ProxEdit \citep{han2024proxedit}, DragDiffusion \citep{shi2024dragdiffusion}, and AnyEdit \citep{yu2025anyedit} highlight the trend toward general-purpose frameworks.  

\noindent \textbf{Scientific Chart Editing.}  
Charts encode structured data, calibrated axes, and embedded text, making editing distinct from natural imagery \citep{brooks2023instructpix2pixlearningfollowimage,huang2024smartedit,han2024proxedit,sheynin2024emu,Brack_2024_CVPR,shi2024dragdiffusion,yu2025anyedit,10884879}. Early efforts include ScImage \citep{zhang2024scimagegoodmultimodallarge}, AutomaTikZ \citep{belouadi2024automatikztextguidedsynthesisscientific}, and ChartEdit \citep{zhao2025chartedit}. However, most pipelines rely on intermediate code (e.g., matplotlib), emphasizing executability but overlooking perceptual quality and downstream usability. This gap motivates benchmarks and methods tailored to figure editing as a distinct research problem.

A more detailed discussion of related work is provided in Appx.~\ref{appendix:related}.

\begin{table*}[t!]
    \centering
    \caption{Comparison of our proposed benchmark  with existing chart-related benchmarks. While prior benchmarks mainly target captioning, QA, or chart-to-code generation, they provide limited coverage of editing operations and interactive settings. 
    \textit{FigEdit} is the first benchmark designed for evaluation of figure editing, supporting diverse chart types, multiple instruction categories, and interactive scenarios such as visual guidance and style transfer edits.
    }
    \label{tab:benchmark_comparison}
    \resizebox{\textwidth}{!}{
    \begin{tabular}{lc|ccccccccc}
        \toprule
        \textbf{Name} & 
        \begin{tabular}{@{}c@{}} \textbf{Output} \\ \textbf{Format} \end{tabular} & 
        \begin{tabular}{@{}c@{}} \textbf{w/ Real} \\ \textbf{Data} \end{tabular} & 
        \begin{tabular}{@{}c@{}} \textbf{Diverse} \\ \textbf{Types} \end{tabular} & 
        \begin{tabular}{@{}c@{}} \textbf{Visual} \\ \textbf{Guidance} \end{tabular} & 
        \begin{tabular}{@{}c@{}} \textbf{Style} \\ \textbf{Transfer} \end{tabular} & 
        \multicolumn{5}{c}{\textbf{Editing Instruction}} \\
        \cmidrule(lr){7-11} 
        & & & & & & \textbf{Data} & \textbf{Format} & \textbf{Layout} & \textbf{Style} & \textbf{Text} \\
        \midrule
        
        ChartCraft \citep{yan2024chartreformer} & Json & \xmark & \xmark & \xmark & \xmark & \cmark & \cmark & \cmark & \cmark & \xmark \\
        Plot2Code \citep{wu2024plot2code} & Code & \cmark & \cmark & \xmark & \xmark & \xmark & \xmark & \xmark & \xmark & \xmark \\
        ChartX \citep{xia2024chartx} & Code & \xmark & \cmark & \xmark & \xmark & \xmark & \xmark & \xmark & \xmark & \xmark \\
        AcademiaChart \citep{zhang2024gpt} & Code & \xmark & \xmark & \xmark & \xmark & \xmark & \xmark & \xmark & \xmark & \xmark \\
        ChartMimic \citep{shi2024chartmimic} & Code & \cmark & \cmark & \xmark & \xmark & \cmark & \xmark & \xmark & \xmark & \xmark \\
        ChartEdit \citep{zhao2025chartedit} & Code & \cmark & \cmark & \xmark & \xmark & \cmark & \cmark & \cmark & \cmark & \cmark \\
        \hline
        FigEdit (Ours) & Figure 
        & \cmark & \cmark & \cmark & \cmark & \cmark & \cmark & \cmark & \cmark & \cmark \\
        \bottomrule
    \end{tabular}
    }
\vspace{-0.5cm}
\end{table*}

\section{Benchmark}
We introduce a \emph{figure–centric} benchmark for scientific figure editing. Ground truth (GT) images are obtained by applying deterministic edit functions to Vega\footnote{https://vega.github.io/}/Vega–Lite\footnote{https://vega.github.io/vega-lite/} specifications and rendering the results. Evaluation is performed in image space. This design provides pixel-consistent supervision across atomic edits, one-shot composite edits, multi-turn conversations, figure edits with visual guidance, and figure edits with referenced figures, without depending on package-specific code.

\begin{table}[t]
\centering
\caption{Benchmark data statistics across chart types and editing tasks. Each entry shows the number of instances per task, with subtotals by chart family and overall totals.
}
\label{tab:chart_type_stats}
\resizebox{\textwidth}{!}{
\begin{tabular}{lrrrrr}
\toprule
\textbf{Chart Type} & \textbf{Single Edit} & \textbf{Multi Edit} & \textbf{Conv. Edit} & \textbf{Style Transfer} & \textbf{Visual Guidance} \\
\midrule
Area         & 1463 & 586 & 369 & 406 & 299 \\
Line         & 1649 & 593 & 398 & 424 & 399 \\
\midrule
Bar          & 1800 & 600 & 375 & 410 & 398 \\
Stacked-bar  & 1800 & 600 & 398 & 498 & 396 \\
\midrule
Pie          & 1000 & 600 & 398 & 200 & 397 \\
Donut        & 1000 & 600 & 399 & 200 & 398 \\
\midrule
\multicolumn{6}{l}{\textbf{Other}} \\
Box          & 1199 & 568 & 388 & 200 & 198 \\
Violin       & 1000   & 500   & 300   & 200   & 150 \\
Scatter      & 1398 & 598 & 324 & 400 & 322 \\
Dot          & 1796 & 599 & 383 & 458 & 398 \\
\midrule
\textbf{Totals}      & \textbf{14105} & \textbf{6244} & \textbf{3732} & \textbf{3400} & \textbf{3355} \\
\multicolumn{6}{r}{\textit{All tasks combined:} \textbf{30836}} \\
\bottomrule
\end{tabular}
}
\vspace{-0.5cm}
\end{table}

\subsection{Formal Definition of a Chart}

A natural image $I$ can be viewed as a function mapping 2D coordinates to color values, $I: \mathbb{R}^2 \rightarrow \mathbb{R}^3$. In contrast, a chart is the rendered output of a structured specification. Formally, we define a deterministic renderer $R$ that maps a specification $\sigma \in \Sigma$ to an image $I \in \mathbb{R}^{H \times W \times 3}$:
\begin{equation}
    I = R(\sigma).
\end{equation}

Each specification $\sigma$ can be decomposed into two components:
\[
\sigma = (C, S),
\]
where Content ($C$) denotes a dataset $D$, a chart type $\tau$, and a mapping function that encodes variables in $D$ to geometric marks. Style ($S$) denotes the visual configuration, including palettes, fonts, strokes/fills, gridlines, legend layout, spacing, and margins.

An atomic edit $e \in \mathcal{E}$ is a total function $f_e:\Sigma \rightarrow \Sigma$, with pre-/post-conditions on $(C,S)$. Given an initial specification $\sigma$ with rendered image $I=R(\sigma)$ and an instruction $u$, a model $M$ produces either an image $\widehat{I}=M(I,u)$ or a specification $\widehat{\sigma}=M(I,u)$.

\subsection{Tasks}

\paragraph{Task 1: Single Chart Edit.}
Given $(I,u)$ where $u$ specifies one atomic edit $e$, the updated specification is as follows:
\[
\sigma^\star = f_e(\sigma), \qquad I^\star = R(\sigma^\star).
\]

\paragraph{Task 2: Multiple Chart Edits.}
Given $(I,u)$ where $u$ specifies $k \geq 2$ atomic edits $\{e_1,\dots,e_k\}$ applied jointly, the updated specification is
\[
\sigma^\star = (f_{e_k} \circ \cdots \circ f_{e_1})(\sigma), \qquad I^\star = R(\sigma^\star).
\]
For non-commutative edits, we adopt a fixed canonical order in the generator.

\paragraph{Task 3: Conversational Chart Edits.}
A session consists of $T$ rounds. At round $t$, the input is $(I_{t-1}, H_{t-1}, u_t)$, where $I_{t-1}$ is the previous image, $H_{t-1}$ is the dialogue history, and $u_t$ is the current instruction. The updated specification is
\[
\sigma_t^\star = (f_{e_t} \circ \cdots \circ f_{e_1})(\sigma), \qquad I_t^\star = R(\sigma_t^\star).
\]

\paragraph{Task 4: Style Transfer.}
Given a source chart $I_s=R(\sigma_s)$ and target content $(D_t, \tau_t)$, the goal is to preserve the target content while adopting the source’s style:
\[
C(\sigma^\star) = (D_t, \tau_t), \qquad S(\sigma^\star) \approx S(\sigma_s), \qquad I^\star = R(\sigma^\star).
\]

\paragraph{Task 5: Visual-Guidance Edits.}
Given $(I, u, \mathcal{G})$, where $\mathcal{G}$ is visual guidance, the goal is to apply the edit $u$ within the guided region while preserving other regions:
\[
\sigma^\star = f_{e, u, \mathcal{G}}(\sigma), \qquad I^\star = R(\sigma^\star).
\]

\subsection{Base Figure Sourcing and Generation}
\label{sec:basefig}
To construct base figures, we define a set of chart classes $\mathcal{C}$ and associate them with curated datasets $\mathcal{A}$ drawn from public sources (full list in Appx.~\ref{appendix:datasets}). 
Each chart class $c \in \mathcal{C}$ is paired with a preference list $\mathcal{P}(c)$ to encourage semantically coherent choices.
We employ a LLM to propose candidate  specifications conditioned on class hints and dataset lists. 
A set of automatic validation and filtering rules ensures that generated charts satisfy schema requirements, avoid duplicates, and maintain semantic diversity. 
In addition, heuristic alignment between dataset domains and chart types further improves quality and coverage. 
All generations are logged with provenance information, and further implementation details are provided in Appx.~\ref{appx:basefig}.

\subsection{Editing Operations}
\label{sec:ops}

We build a suite of editing tasks derived from a canonical operation set $\mathcal{O}$ (See Appx.~\ref{sec:ops_appendix} for more details). 
Each element in $\mathcal{O}$ encodes an atomic edit, covering text, style, layout, and data–centric manipulations. 
Invalid operations are filtered out depending on chart semantics (e.g., spacing edits require band/point scales).

From each chart we automatically produce (i) natural–language instructions augmented with machine–readable OP tags\footnote{OP = operation; each OP tag encodes the intended atomic edit.}, 
(ii) edited specifications with inline data values, and (iii) corresponding rendered images. 
On top of these atomic edits, we derive (iv) conversational annotations that align multi–step edits with their constituent single edits, 
(v) visual–guidance assets where the target region is circled on the original chart, 
and (vi) style–transfer annotations that pair a target edit with a reference figure providing the desired style attribute. 
More details are provided in Appx.~\ref{sec:ops_appendix}.

\subsubsection{Single and Multi Edit Generation}
\label{sec:ops:gen}
For each chart we sample a feasible subset $\mathcal{O}(c)\subseteq \mathcal{O}$ 
and realize the edits as natural instructions with corresponding OP tags. 
Edited specifications are validated to preserve schema correctness, ensure visible changes, 
and maintain consistent data accounting when adding or removing rows. 
These checks guarantee deterministic and reproducible supervision.

\subsubsection{Conversational Annotations}
\label{sec:ops:combo2}
We further construct short multi–turn conversations by decomposing a two–step edit into its constituent single edits. Each conversational sample provides the original chart, two turns of instructions with their intermediate ground truth states, and the final outcome. This setting evaluates whether models can maintain state and history across turns rather than only executing isolated edits.

\subsubsection{Visual–Guidance Assets}
\label{sec:ops:visual}
For a selected subset of operations, 
we create visually grounded variants by marking the target region directly on the original chart. 
To generate the visual overlay, we employ a vision–language model (GPT-Image) that is prompted to draw a thin red circle around the specified element while leaving chart content unchanged. 
Each sample provides both a concise natural instruction and a guidance image with the circled target. 
This variant enables evaluation of multimodal understanding, where the model must integrate textual instructions with explicit visual cues.

\subsubsection{Style–Transfer Annotations}
\label{sec:ops:transfer}
Finally, we introduce a style–transfer setting in which an edited chart is paired with a reference chart whose current style attribute matches the target of the edit. The model is asked to reproduce the target chart while adopting the style of the reference. This task connects editing with cross–figure style adaptation and highlights the challenge of disentangling content from stylistic attributes.


\begin{table}[t]
\centering
\caption{Distribution of editing operations by task. Operations are grouped into categories such as data-centric, text, style, and layout, with counts reported per task and overall totals.}
\label{tab:ops_summary_all}
\resizebox{\textwidth}{!}{
\begin{tabularx}{\textwidth}{l l X r}
\toprule
\textbf{Task} & \textbf{Category} & \textbf{Operation} & \textbf{Image Count} \\
\hline
\multirow{7}{*}{\textbf{Single Edit}} & \multirow{2}{*}{Data-centric} & Add element & 1941 \\
 &  & Remove element & 1892 \\
 & \multirow{1}{*}{Text} & Add title & 1942 \\
 & \multirow{2}{*}{Style Editing} & Change background color & 1944 \\
 &  & Change data color & 1729 \\
 & \multirow{1}{*}{Margin Adjustments} & Adjust category spacing & 1729 \\
 & \multirow{1}{*}{Font} & Font Adjustment & 2943 \\
\hline
\multirow{2}{*}{\textbf{Multi Edit}} & \multirow{1}{*}{Dual-operation} & Combine 2 edits & 3370 \\
 & \multirow{1}{*}{Triple-operation} & Combine 3+ edits & 2660 \\
\hline
\multirow{1}{*}{\textbf{Conversational Edit}} & \multirow{1}{*}{} &  & 3575 \\
\hline
\multirow{2}{*}{\textbf{Visual Guidance}} & \multirow{1}{*}{Style Editing} & Change data color & 1666 \\
 & \multirow{1}{*}{Data-centric} & Remove element & 1819 \\
\hline
\multirow{3}{*}{\textbf{Style Transfer}} & \multirow{1}{*}{Style Mapping} & Transfer style & 1511 \\
 & \multirow{1}{*}{Style Editing} & Change data color & 1728 \\
 & \multirow{1}{*}{Margin Adjustments} & Adjust category spacing & 387 \\
\hline
\multicolumn{3}{r}{\textbf{Overall Total}} & \textbf{30836} \\
\bottomrule
\end{tabularx}
}
\vspace{-0.5cm}
\end{table}

\begin{figure*}[t]
  \centering
  \includegraphics[width=\textwidth]{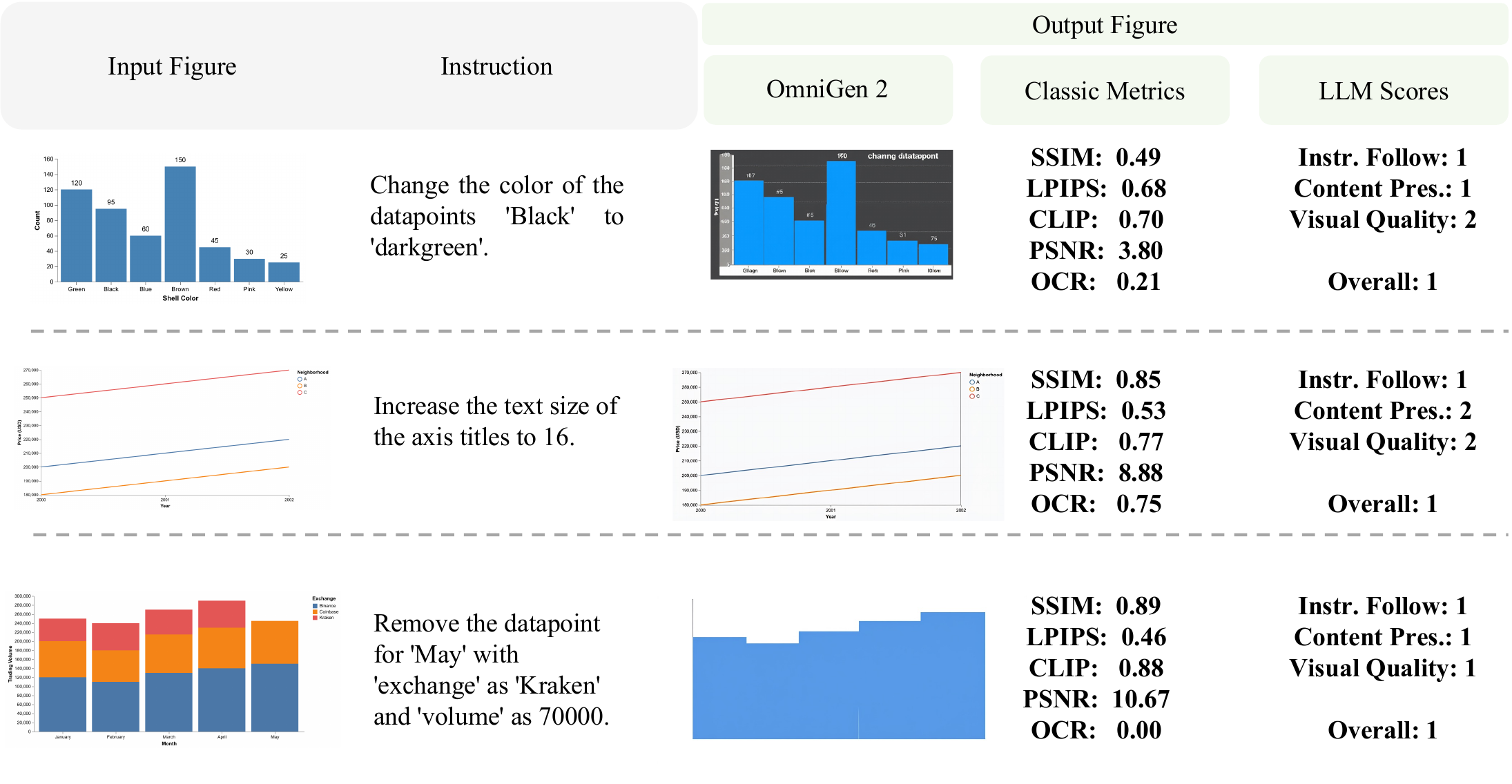}
  \caption{
  Comparison of chart editing evaluation signals on three representative cases. 
  The left block shows the \emph{Input Figure} and the \emph{Instruction}. 
  The right block shows the \emph{Output Figure} from OmniGen2, the \emph{Classic Metrics} (e.g., SSIM and PSNR), and the \emph{LLM Scores}.
  We observe that classic pixel metrics can remain high while the edit is wrong.
  This reveals a gap between pixel similarity and semantic edit correctness, which motivates semantics-aware evaluation for figure editing.
  }
  \label{fig:metric-gap}
\vspace{-0.3cm}
\end{figure*}

\subsection{Dataset Statistics}
\label{sec:stat}
The final benchmark contains 30,836 edited figures, distributed across five task families. 
Tab.~\ref{tab:ops_summary_all} summarizes the counts by operation type. 
Single edits form the largest portion of the dataset, covering basic manipulations such as element addition/removal, text and font changes, color and background modifications, and spacing adjustments, totaling 14,105 figures. 
Multi edits contribute another 6,244 examples, split between dual edits and three–operation combinations. 
Conversational settings add 3,732 two–turn sequences, while the visual–guidance and style–transfer tasks contribute 3,355 and 3,400 figures, respectively. 
Together, these distributions provide a balanced coverage of atomic edits, composite edits, multimodal guidance, and cross–style adaptation.
A breakdown by chart type is shown in Tab.~\ref{tab:chart_type_stats}. 
Importantly, all base figures are derived from \emph{real-world datasets}, spanning domains such as economics, climate, healthcare, sports, and social science. 
A complete list of datasets used in figure generation is provided in Appx.~\ref{appendix:datasets}.

\begin{figure}[t]
\centering
\includegraphics[width=0.95\textwidth]{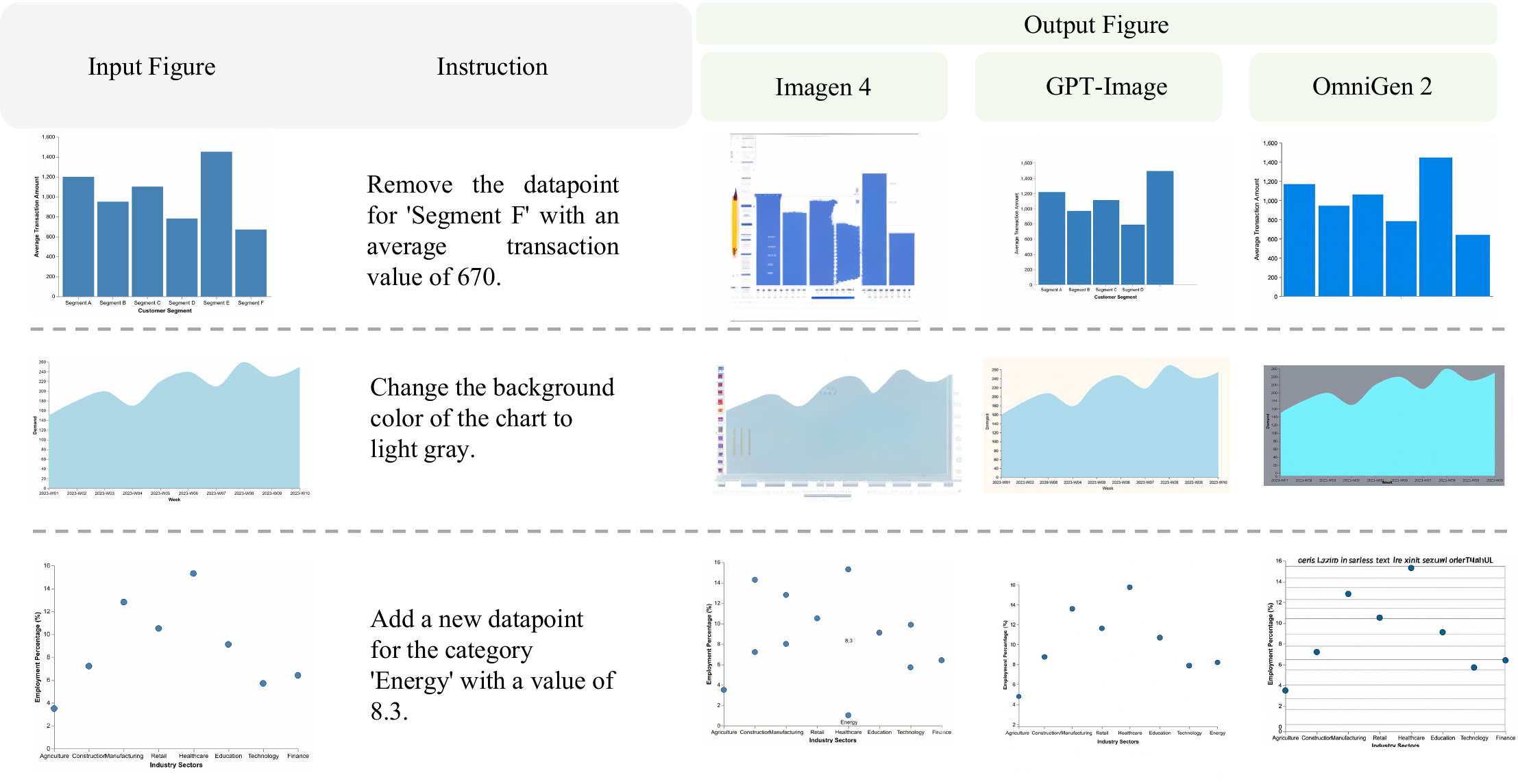} 
\vspace{-0.1in}
   \caption{Qualitative examples of figure editing with three representative instructions. For each case, the input figure and target instruction are shown on the left, and outputs from Imagen 4, GPT-Image, and OmniGen2 are shown on the right.}
\label{fig:qualitative}
\vspace{-0.3cm}
\end{figure}

\begin{figure*}[t]
\centering
\begin{subfigure}{0.48\linewidth}
    \centering
    \includegraphics[width=0.98\linewidth]{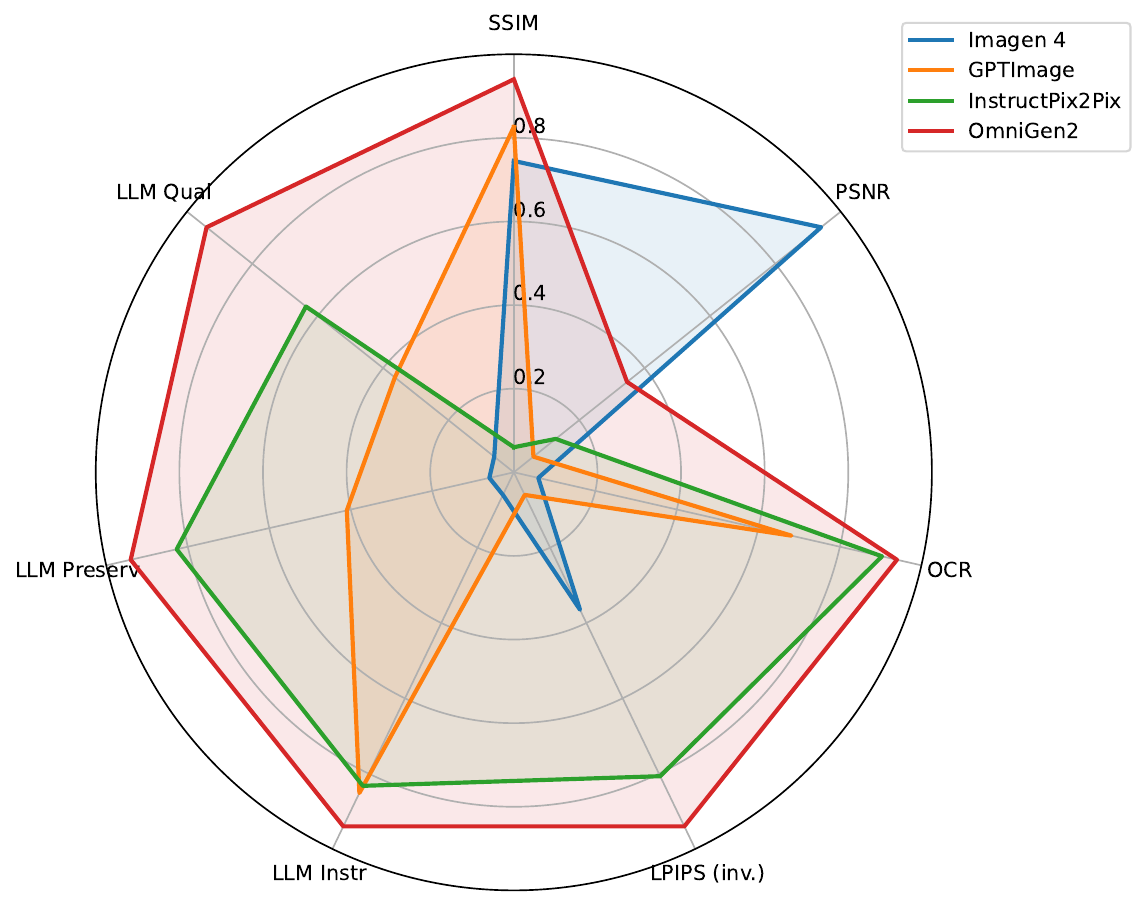}
    \caption{Multi Edit Task}
    \label{fig:radar_single}
\end{subfigure}
\hfill
\begin{subfigure}{0.48\linewidth}
    \centering
    \includegraphics[width=0.98\linewidth]{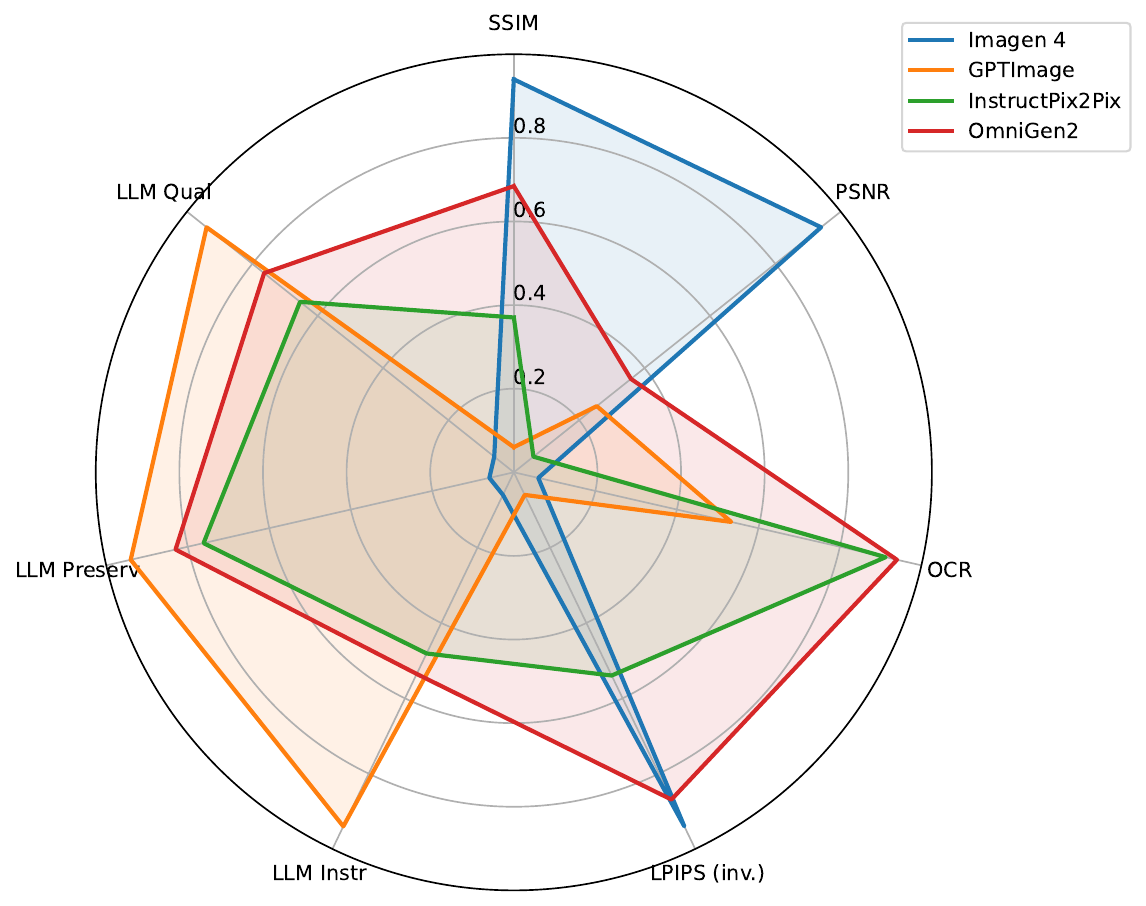}
    \caption{Conversational Edit Task}
    \label{fig:radar_conv}
\end{subfigure}
\caption{Radar charts for different tasks (normalized with epsilon, LPIPS inverted). 
Each chart compares all models on SSIM, PSNR, OCR, LPIPS, and three LLM scores.}
\label{fig:radar_multi_conv}
\vspace{-0.5cm}
\end{figure*}

\subsection{Evaluation Protocol}
\label{sec:evaluation}

We evaluate all models directly in image space. We compute six complementary metrics: SSIM~\citep{wang2004ssim}, PSNR~\citep{hore2010psnr}, LPIPS~\citep{zhang2018lpips}, CLIP similarity~\citep{radford2021learning}, OCR similarity~\citep{smith2007tesseract}, and an LLM-based instruction score. The first five are classic metrics widely used in image generation and vision tasks, while the last directly evaluates whether edits satisfy the instruction, preserve chart content, and maintain visual quality. More details on implementations are provided in Appx.~\ref{app:metrics}.

\begin{table*}[t]
\centering
\caption{Performance comparison grouped by task. Higher is better for SSIM, CLIP, PSNR, OCR, and LLM Scores. Lower is better for LPIPS. \textbf{Instr.} denotes instruction following score. \textbf{Preserv.} denotes content preservation score. \textbf{Qual.} denotes image quality score.}
\label{tab:task_metrics}
\footnotesize
\setlength{\tabcolsep}{4pt}
\begin{tabular}{llccccccccc}
\toprule
\textbf{Task} & \textbf{Model} & \textbf{SSIM} $\uparrow$ & \textbf{LPIPS} $\downarrow$ & \textbf{CLIP} $\uparrow$ & \textbf{PSNR} $\uparrow$ & \textbf{OCR} $\uparrow$ & \multicolumn{3}{c}{\textbf{MLLM Score (1--5)} $\uparrow$} \\
\cmidrule(lr){8-10}
& & & & & & & \textbf{Instr.} & \textbf{Preserv.} & \textbf{Qual.} \\
\midrule
\multirow{4}{*}{\textbf{Single}} 
& Imagen 4         & \textbf{0.7726} & \textbf{0.4094} & 0.7781 & \textbf{13.04} & 0.0723 & 1.58 & 1.51 & 2.05 \\
& GPTImage         & 0.7295 & 0.5383 & 0.8099 & 10.32 & 0.2054 & \textbf{3.47} & 1.71 & 2.45 \\
& InstructPix2Pix  & 0.7211 & 0.4811 & 0.8328 & 11.02 & 0.2568 & 3.27 & 2.50 & 2.77 \\
& OmniGen2         & 0.7350 & 0.4705 & \textbf{0.8350} & 11.30 & \textbf{0.2620} & 3.35 & \textbf{2.55} & \textbf{2.85} \\
\midrule
\multirow{4}{*}{\textbf{Multi}} 
& Imagen 4         & 0.6958 & 0.5549 & 0.7738 & \textbf{11.02} & 0.1069 & 1.26 & 1.32 & 2.15 \\
& GPTImage         & 0.7017 & 0.5787 & \textbf{0.8070} &  9.73 & 0.2185 & 2.51 & 1.63 & 2.34 \\
& InstructPix2Pix  & 0.6460 & 0.5204 & 0.8043 &  9.83 & 0.2584 & 2.48 & 2.00 & 2.51 \\
& OmniGen2         & \textbf{0.7100} & \textbf{0.5100} & 0.8220 & 10.15 & \textbf{0.2650} & \textbf{2.65} & \textbf{2.10} & \textbf{2.70} \\
\midrule
\multirow{4}{*}{\textbf{Conv.}} 
& Imagen 4         & \textbf{0.7180} & \textbf{0.4923} & 0.7599 & \textbf{11.58} & 0.0698 & 1.35 & 1.23 & 2.11 \\
& GPTImage         & 0.6732 & 0.5257 & \textbf{0.8525} & 10.66 & 0.1721 & \textbf{4.59} & \textbf{2.51} & \textbf{2.91} \\
& InstructPix2Pix  & 0.6890 & 0.5075 & 0.8200 & 10.40 & 0.2540 & 2.90 & 2.25 & 2.65 \\
& OmniGen2         & 0.7050 & 0.4950 & 0.8280 & 10.80 & \textbf{0.2600} & 3.10 & 2.35 & 2.75 \\
\midrule
\multirow{4}{*}{\textbf{Visual}} 
& Imagen 4         & \textbf{0.8420} & \textbf{0.5050} & 0.7600 & \textbf{13.10} & 0.1200 & 1.40 & 1.35 & 2.20 \\
& GPTImage         & 0.8355 & 0.5207 & \textbf{0.8444} & 12.85 & \textbf{0.4665} & \textbf{2.39} & \textbf{3.16} & \textbf{3.95} \\
& InstructPix2Pix  & 0.7380 & 0.5220 & 0.8190 & 10.90 & 0.2200 & 1.85 & 2.20 & 2.80 \\
& OmniGen2         & 0.7508 & 0.5236 & 0.8187 &  8.98 & 0.1806 & 1.19 & 1.85 & 2.74 \\
\midrule
\multirow{4}{*}{\textbf{Transfer}} 
& Imagen 4         & \textbf{0.8500} & 0.4800 & 0.7700 & \textbf{14.00} & 0.1300 & 1.30 & 1.25 & 2.15 \\
& GPTImage         & 0.8438 & 0.4934 & 0.8054 & 13.81 & \textbf{0.5092} & \textbf{3.06} & \textbf{3.57} & \textbf{4.16} \\
& InstructPix2Pix  & 0.7960 & 0.5020 & \textbf{0.8160} & 12.90 & 0.2400 & 2.20 & 2.60 & 3.10 \\
& OmniGen2         & 0.8246 & \textbf{0.4376} & 0.8127 & 12.08 & 0.3147 & 1.53 & 2.14 & 2.64 \\
\bottomrule
\end{tabular}
\end{table*}

\section{Experiment}

\noindent\textbf{Baselines.} 
We evaluate against four representative instruction-based editing models: GPT-Image~\citep{OpenAI_GPTImage}, Imagen~4~\citep{GoogleImagen4_2025}, OmniGen~2~\citep{wu2025omnigen2}, and InstructPix2Pix~\citep{brooksinst}. 
These span closed–source commercial systems and open–source research frameworks, covering both diffusion-based editors and multimodal approaches. 
Further details on each baseline are provided in Appx.~\ref{app:baselines}.

\noindent \textbf{Experiment Setup.}
We evaluate chart editing across five tasks.
All methods operate on the same set of instructions and images. 
Prompts are standardized to encourage strictly local modifications while maintaining axes, labels, and other contextual elements. 
Further implementation details are provided in Appx.~\ref{appx:exp-details}.

\subsection{Main Results}
\label{sec:mainres}
\noindent \textbf{Overall performance across tasks.}
Tab.~\ref{tab:task_metrics} summarizes the performance of representative editing models across the five evaluation settings. 
Imagen 4 achieves consistently high scores on SSIM and PSNR, reflecting strong pixel-level resemblance to the input figures, but its instruction-following and preservation scores are the lowest among all models. 
GPT-Image excels in conversational and transfer settings, showing the highest instruction-following scores, but often sacrifices content fidelity. 
OmniGen2 strikes a balance, performing reliably across most tasks with solid LLM scores and relatively stable OCR accuracy. 
InstructPix2Pix remains competitive but generally underperforms OmniGen2, particularly on complex edits, while still clearly surpassing Imagen 4 on semantic alignment. 
These results highlight that strong performance on pixel-based similarity metrics does not necessarily translate into correct or faithful edits.

\paragraph{Limitations of classic metrics.}
Fig.~\ref{fig:radar_multi_conv} provides a more detailed comparison of multi-edit and conversational tasks. 
Classic metrics such as SSIM and PSNR exaggerate the performance of pixel-oriented models like Imagen 4, while LLM-based scores and OCR accuracy reveal significant semantic errors. 
The radar plots make this gap visually explicit: models that appear strong under pixel similarity collapse when judged by whether the requested edits were actually applied. 
This finding is consistent with the qualitative evidence in Fig.~\ref{fig:qualitative} and reinforces the need for evaluation protocols that go beyond pixel resemblance.

\paragraph{Per-instruction breakdown.}
We further analyze performance at the level of individual instructions in Appx.~\ref{appx:more_res}.
These results confirm the same trend: models often achieve high SSIM or PSNR even when edits such as adding datapoints or changing axis labels are not correctly applied.

\subsection{Analysis}

\noindent \textbf{The gap between pixel-level similarity and semantic correctness.}
Fig.~\ref{fig:metric-gap} and Tab.~\ref{tab:task_metrics} highlight a consistent limitation of classic image metrics in the context of figure editing. 
Models such as Imagen 4 and OmniGen2 can obtain high SSIM and PSNR scores, yet their outputs often fail to apply the intended transformation. 
As illustrated in Fig.~\ref{fig:metric-gap}, edits may preserve overall appearance while the instruction is ignored, the figure is distorted, or key content is changed. 
Tab.~\ref{tab:task_metrics} shows the same pattern across tasks: pixel-based metrics remain strong, but instruction-following and content-preservation scores from LLM-based evaluation drop sharply, especially for multi-step and conversational edits. 
These results indicate that similarity at the pixel level is not a reliable indicator of semantic correctness. 
They also motivate the need for benchmarks that evaluate edits at the level of data and visual encodings rather than image resemblance alone.

\noindent \textbf{No single model dominates across tasks.}
Tab.~\ref{tab:task_metrics} shows that performance is highly fragmented: no model achieves consistently strong results across all task types or metrics. Imagen~4 tends to lead on low-level pixel fidelity metrics such as SSIM and PSNR, yet it performs poorly on instruction-following and semantic preservation, indicating that its edits often look visually smooth but fail to reflect the requested change. GPT-Image shows the opposite trend: it excels in instruction scores, especially in conversational and transfer settings, but lags behind on PSNR and OCR accuracy, suggesting weaker robustness to text-heavy or layout-sensitive edits. InstructPix2Pix performs competitively on some semantic metrics but is generally less reliable than OmniGen2, which offers a more balanced profile. However, OmniGen2 also struggles with visual-guided and transfer edits, highlighting its limitations in cross-instance reasoning. These results reveal that current models overfit to specific task structures or metric types, and that strong performance on classic pixel-level metrics does not guarantee reliable edit satisfaction in more challenging scenarios.

\noindent \textbf{Qualitative study.}
Fig.~\ref{fig:qualitative} illustrates representative failure cases in figure editing. 
Across different instructions: removing a datapoint, changing a background color, or adding a new element, current models frequently produce outputs that appear visually similar yet fail to realize the requested transformation. 
These cases mirror the quantitative results: classic pixel-level metrics often remain high even when semantic correctness is violated. 
The examples highlight how generative editors, optimized for perceptual similarity, struggle with structure-preserving transformations, reinforcing the need for evaluation protocols and benchmarks that explicitly target semantic consistency.
More cases can be found in Appx.~\ref{fig:add_qualitative}.

\section{Conclusion}
\label{sec:conclusion}
We introduced \textit{FigEdit}, a large-scale benchmark for scientific figure editing that treats editing as a structured transformation problem grounded in graphical grammar. The benchmark spans diverse chart types and task settings, and it provides both figure outputs and executable specifications to support reliable evaluation. Our experiments show that existing models perform poorly when edits require semantic consistency, which reveals a clear gap between current approaches and the needs of figure editing. By offering a task-structured and semantics-aware evaluation protocol, \textit{FigEdit} establishes a foundation for developing future models that can perform faithful, data-aligned, and auditable edits.

\section{Ethics Statement \& Reproducibility Statement}
This work adheres to standard academic research practices. 
All data used are either publicly available or synthetically generated, and the study is intended solely for scientific and educational purposes. 
We do not foresee any ethical concerns arising from the content or methodology presented. 
For reproducibility, we have included sufficient technical details in the paper to allow other researchers to replicate our experiments. 
The dataset statistics, task definitions, and evaluation protocols are described in detail, and we aim to facilitate further exploration and extension by the community.  

\bibliography{iclr2023_conference}

@inproceedings{li2024enhanced,
  title={Enhanced Visual Instruction Tuning with Synthesized Image-Dialogue Data},
  author={Li, Yanda and Zhang, Chi and Yu, Gang and Yang, Wanqi and Wang, Zhibin and Fu, Bin and Lin, Guosheng and Shen, Chunhua and Chen, Ling and Wei, Yunchao},
  booktitle={Findings of the Association for Computational Linguistics ACL 2024},
  pages={14512--14531},
  year={2024}
}

@inproceedings{ma2025dialogdraw,
  title={DialogDraw: Image Generation and Editing System Based on Multi-Turn Dialogue},
  author={Ma, Shichao and Zhang, Xinfeng and Zhao, Zeng and Liu, Bai and Fan, Changjie and Hu, Zhipeng},
  booktitle={Proceedings of the AAAI Conference on Artificial Intelligence},
  volume={39},
  number={23},
  pages={24795--24803},
  year={2025}
}

@article{taneja2025mudoc,
  title={MuDoC: An Interactive Multimodal Document-grounded Conversational AI System},
  author={Taneja, Karan and Goel, Ashok K},
  journal={arXiv preprint arXiv:2502.09843},
  year={2025}
}

@inproceedings{wei2024balancing,
  title={Balancing visual context understanding in dialogue for image retrieval},
  author={Wei, Zhaohui and Liao, Lizi and Du, Xiaoyu and Xiang, Xinguang},
  year={2024},
  organization={Association for Computational Linguistics}
}

@article{huang2024dialoggen,
  title={Dialoggen: Multi-modal interactive dialogue system for multi-turn text-to-image generation},
  author={Huang, Minbin and Long, Yanxin and Deng, Xinchi and Chu, Ruihang and Xiong, Jiangfeng and Liang, Xiaodan and Cheng, Hong and Lu, Qinglin and Liu, Wei},
  journal={arXiv preprint arXiv:2403.08857},
  year={2024}
}

@article{liu2024you,
  title={What Do You Want? User-centric Prompt Generation for Text-to-image Synthesis via Multi-turn Guidance},
  author={Liu, Yilun and He, Minggui and Yao, Feiyu and Ji, Yuhe and Tao, Shimin and Du, Jingzhou and Li, Duan and Gao, Jian and Zhang, Li and Yang, Hao and others},
  journal={arXiv preprint arXiv:2408.12910},
  year={2024}
}

@article{zhao2025chatsearch,
  title={Chatsearch: A dataset and a generative retrieval model for general conversational image retrieval},
  author={Zhao, Zijia and Guo, Longteng and Yue, Tongtian and Hu, Erdong and Shao, Shuai and Yuan, Zehuan and Huang, Hua and Liu, Jing},
  journal={Pattern Recognition},
  pages={111696},
  year={2025},
  publisher={Elsevier}
}

@article{zhao2025chartedit,
  title={ChartEdit: How Far Are MLLMs From Automating Chart Analysis? Evaluating MLLMs' Capability via Chart Editing},
  author={Zhao, Xuanle and Liu, Xuexin and Yang, Haoyue and Luo, Xianzhen and Zeng, Fanhu and Li, Jianling and Shi, Qi and Chen, Chi},
  journal={arXiv preprint arXiv:2505.11935},
  year={2025}
}

@article{wang2025twin,
title={Twin Co-Adaptive Dialogue for Progressive Image Generation},
author={Wang, Jianhui and He, Yangfan and Zhong, Yan and Song, Xinyuan and Su, Jiayi and Feng, Yuheng and He, Hongyang and Zhu, Wenyu and Yuan, Xinhang and Lu, Kuan and others},
journal={arXiv preprint arXiv:2504.14868},
year={2025}
}

@article{hahn2024proactive,
title={Proactive Agents for Multi-Turn Text-to-Image Generation Under Uncertainty},
author={Hahn, Meera and Zeng, Wenjun and Kannen, Nithish and Galt, Rich and Badola, Kartikeya and Kim, Been and Wang, Zi},
journal={arXiv preprint arXiv:2412.06771},
year={2024}
}

@article{chen2024learning,
title={Learning to Clarify: Multi-turn Conversations with Action-Based Contrastive Self-Training},
author={Chen, Maximillian and Sun, Ruoxi and Ar{\i}k, Sercan {\"O} and Pfister, Tomas},
journal={arXiv preprint arXiv:2406.00222},
year={2024}
}

@article{deng2025proactive,
title={Proactive Conversational AI: A Comprehensive Survey of Advancements and Opportunities},
author={Deng, Yang and Liao, Lizi and Lei, Wenqiang and Yang, Grace and Lam, Wai and Chua, Tat-Seng},
journal={ACM Transactions on Information Systems},
year={2025},
publisher={ACM New York, NY}
}

@article{zhang2025survey,
title={A Survey on Multi-Turn Interaction Capabilities of Large Language Models},
author={Zhang, Chen and Dai, Xinyi and Wu, Yaxiong and Yang, Qu and Wang, Yasheng and Tang, Ruiming and Liu, Yong},
journal={arXiv preprint arXiv:2501.09959},
year={2025}
}

@article{andukuri2024star,
title={Star-gate: Teaching language models to ask clarifying questions},
author={Andukuri, Chinmaya and Fr{\"a}nken, Jan-Philipp and Gerstenberg, Tobias and Goodman, Noah D},
journal={arXiv preprint arXiv:2403.19154},
year={2024}
}

@inproceedings{zelikman2024star,
title={STaR: Self-taught reasoner bootstrapping reasoning with reasoning},
author={Zelikman, Eric and Wu, YH and Mu, Jesse and Goodman, Noah D},
booktitle={Proc. the 36th International Conference on Neural Information Processing Systems},
volume={1126},
year={2024}
}

@misc{ramesh2022hierarchicaltextconditionalimagegeneration,
      title={Hierarchical Text-Conditional Image Generation with CLIP Latents}, 
      author={Aditya Ramesh and Prafulla Dhariwal and Alex Nichol and Casey Chu and Mark Chen},
      year={2022},
      eprint={2204.06125},
      archivePrefix={arXiv},
 
}

@misc{rombach2022highresolutionimagesynthesislatent,
      title={High-Resolution Image Synthesis with Latent Diffusion Models}, 
      author={Robin Rombach and Andreas Blattmann and Dominik Lorenz and Patrick Esser and Björn Ommer},
      year={2022},
      eprint={2112.10752},
      archivePrefix={arXiv}, 
}

@misc{zhang2023addingconditionalcontroltexttoimage,
      title={Adding Conditional Control to Text-to-Image Diffusion Models}, 
      author={Lvmin Zhang and Anyi Rao and Maneesh Agrawala},
      year={2023},
      eprint={2302.05543},
      archivePrefix={arXiv}, 
}

@misc{brooks2023instructpix2pixlearningfollowimage,
      title={InstructPix2Pix: Learning to Follow Image Editing Instructions}, 
      author={Tim Brooks and Aleksander Holynski and Alexei A. Efros},
      year={2023},
      eprint={2211.09800},
      archivePrefix={arXiv},
    
}

@inproceedings{NEURIPS2023_f8ad010c,
 author = {Huang, Kaiyi and Sun, Kaiyue and Xie, Enze and Li, Zhenguo and Liu, Xihui},
 booktitle = {Advances in Neural Information Processing Systems},
 pages = {78723--78747},
 publisher = {Curran Associates, Inc.},
 title = {T2I-CompBench: A Comprehensive Benchmark for Open-world Compositional Text-to-image Generation},
 volume = {36},
 year = {2023}
}

@inproceedings{Wang_2024, series={CHI ’24},
   title={PromptCharm: Text-to-Image Generation through Multi-modal Prompting and Refinement},
   DOI={10.1145/3613904.3642803},
   booktitle={Proceedings of the CHI Conference on Human Factors in Computing Systems},
   author={Wang, Zhijie and Huang, Yuheng and Song, Da and Ma, Lei and Zhang, Tianyi},
   year={2024},
   month=may, pages={1–21},
   collection={CHI ’24} }

@misc{schuhmann2022laion5bopenlargescaledataset,
      title={LAION-5B: An open large-scale dataset for training next generation image-text models}, 
      author={Christoph Schuhmann and Romain Beaumont and Richard Vencu and Cade Gordon and Ross Wightman and Mehdi Cherti and Theo Coombes and Aarush Katta and Clayton Mullis and Mitchell Wortsman and Patrick Schramowski and Srivatsa Kundurthy and Katherine Crowson and Ludwig Schmidt and Robert Kaczmarczyk and Jenia Jitsev},
      year={2022},
      eprint={2210.08402},
      archivePrefix={arXiv},
     
}

@INPROCEEDINGS{10030860,
  author={Rodriguez, Juan A. and Vazquez, David and Laradji, Issam and Pedersoli, Marco and Rodriguez, Pau},
  booktitle={2023 IEEE/CVF Winter Conference on Applications of Computer Vision (WACV)}, 
  title={OCR-VQGAN: Taming Text-within-Image Generation}, 
  year={2023},
  volume={},
  number={},
  pages={3678-3687},
  doi={10.1109/WACV56688.2023.00368}}

@misc{zhang2024scimagegoodmultimodallarge,
      title={ScImage: How Good Are Multimodal Large Language Models at Scientific Text-to-Image Generation?}, 
      author={Leixin Zhang and Steffen Eger and Yinjie Cheng and Weihe Zhai and Jonas Belouadi and Christoph Leiter and Simone Paolo Ponzetto and Fahimeh Moafian and Zhixue Zhao},
      year={2024},
      eprint={2412.02368},
      archivePrefix={arXiv},
    
}

@inproceedings{rodriguez2023ocr,
  title={Ocr-vqgan: Taming text-within-image generation},
  author={Rodriguez, Juan A and Vazquez, David and Laradji, Issam and Pedersoli, Marco and Rodriguez, Pau},
  booktitle={Proceedings of the IEEE/CVF winter conference on applications of computer vision},
  pages={3689--3698},
  year={2023}
}

@misc{belouadi2024automatikztextguidedsynthesisscientific,
      title={AutomaTikZ: Text-Guided Synthesis of Scientific Vector Graphics with TikZ}, 
      author={Jonas Belouadi and Anne Lauscher and Steffen Eger},
      year={2024},
      eprint={2310.00367},
      archivePrefix={arXiv},
    
}

@article{han2023chartllama,
  title={Chartllama: A multimodal llm for chart understanding and generation},
  author={Han, Yucheng and Zhang, Chi and Chen, Xin and Yang, Xu and Wang, Zhibin and Yu, Gang and Fu, Bin and Zhang, Hanwang},
  journal={arXiv preprint arXiv:2311.16483},
  year={2023}
}

@article{masry2023unichart,
  title={Unichart: A universal vision-language pretrained model for chart comprehension and reasoning},
  author={Masry, Ahmed and Kavehzadeh, Parsa and Do, Xuan Long and Hoque, Enamul and Joty, Shafiq},
  journal={arXiv preprint arXiv:2305.14761},
  year={2023}
}

@article{kantharaj2022chart,
  title={Chart-to-text: A large-scale benchmark for chart summarization},
  author={Kantharaj, Shankar and Leong, Rixie Tiffany Ko and Lin, Xiang and Masry, Ahmed and Thakkar, Megh and Hoque, Enamul and Joty, Shafiq},
  journal={arXiv preprint arXiv:2203.06486},
  year={2022}
}

@article{hsu2021scicap,
  title={SciCap: Generating captions for scientific figures},
  author={Hsu, Ting-Yao and Giles, C Lee and Huang, Ting-Hao'Kenneth'},
  journal={arXiv preprint arXiv:2110.11624},
  year={2021}
}

@misc{liu2024llavanext,
    title={LLaVA-NeXT: Improved reasoning, OCR, and world knowledge},
    author={Liu, Haotian and Li, Chunyuan and Li, Yuheng and Li, Bo and Zhang, Yuanhan and Shen, Sheng and Lee, Yong Jae},
    month={January},
    year={2024}
}

@article{guo2024deepseek,
  title={DeepSeek-Coder: When the Large Language Model Meets Programming--The Rise of Code Intelligence},
  author={Guo, Daya and Zhu, Qihao and Yang, Dejian and Xie, Zhenda and Dong, Kai and Zhang, Wentao and Chen, Guanting and Bi, Xiao and Wu, Yu and Li, YK and others},
  journal={arXiv preprint arXiv:2401.14196},
  year={2024}
}

@inproceedings{radford2021learning,
  title={Learning transferable visual models from natural language supervision},
  author={Radford, Alec and Kim, Jong Wook and Hallacy, Chris and Ramesh, Aditya and Goh, Gabriel and Agarwal, Sandhini and Sastry, Girish and Askell, Amanda and Mishkin, Pamela and Clark, Jack and others},
  booktitle={International conference on machine learning},
  pages={8748--8763},
  year={2021},
  organization={PMLR}
}

@inproceedings{chen2024internvl,
  title={Internvl: Scaling up vision foundation models and aligning for generic visual-linguistic tasks},
  author={Chen, Zhe and Wu, Jiannan and Wang, Wenhai and Su, Weijie and Chen, Guo and Xing, Sen and Zhong, Muyan and Zhang, Qinglong and Zhu, Xizhou and Lu, Lewei and others},
  booktitle={Proceedings of the IEEE/CVF Conference on Computer Vision and Pattern Recognition},
  pages={24185--24198},
  year={2024}
}

@article{wang2024qwen2,
  title={Qwen2-vl: Enhancing vision-language model's perception of the world at any resolution},
  author={Wang, Peng and Bai, Shuai and Tan, Sinan and Wang, Shijie and Fan, Zhihao and Bai, Jinze and Chen, Keqin and Liu, Xuejing and Wang, Jialin and Ge, Wenbin and others},
  journal={arXiv preprint arXiv:2409.12191},
  year={2024}
}

@article{lu2024deepseek,
  title={Deepseek-vl: towards real-world vision-language understanding},
  author={Lu, Haoyu and Liu, Wen and Zhang, Bo and Wang, Bingxuan and Dong, Kai and Liu, Bo and Sun, Jingxiang and Ren, Tongzheng and Li, Zhuoshu and Yang, Hao and others},
  journal={arXiv preprint arXiv:2403.05525},
  year={2024}
}

@article{li2024llava,
  title={LLaVA-NeXT-Interleave: Tackling Multi-image, Video, and 3D in Large Multimodal Models},
  author={Li, Feng and Zhang, Renrui and Zhang, Hao and Zhang, Yuanhan and Li, Bo and Li, Wei and Ma, Zejun and Li, Chunyuan},
  journal={arXiv preprint arXiv:2407.07895},
  year={2024}
}

@article{xu2024llava,
  title={Llava-uhd: an lmm perceiving any aspect ratio and high-resolution images},
  author={Xu, Ruyi and Yao, Yuan and Guo, Zonghao and Cui, Junbo and Ni, Zanlin and Ge, Chunjiang and Chua, Tat-Seng and Liu, Zhiyuan and Sun, Maosong and Huang, Gao},
  journal={arXiv preprint arXiv:2403.11703},
  year={2024}
}

@article{yao2024minicpm,
  title={Minicpm-v: A gpt-4v level mllm on your phone},
  author={Yao, Yuan and Yu, Tianyu and Zhang, Ao and Wang, Chongyi and Cui, Junbo and Zhu, Hongji and Cai, Tianchi and Li, Haoyu and Zhao, Weilin and He, Zhihui and others},
  journal={arXiv preprint arXiv:2408.01800},
  year={2024}
}

@article{team2023gemini,
  title={Gemini: a family of highly capable multimodal models},
  author={Team, Gemini and Anil, Rohan and Borgeaud, Sebastian and Alayrac, Jean-Baptiste and Yu, Jiahui and Soricut, Radu and Schalkwyk, Johan and Dai, Andrew M and Hauth, Anja and Millican, Katie and others},
  journal={arXiv preprint arXiv:2312.11805},
  year={2023}
}

@misc{openai2024gpt4o,
  author       = {OpenAI},
  title        = {GPT-4o},
  year         = {2024},
  url          = {https://openai.com/index/hello-gpt-4o},
  note         = {Accessed: 2024-05-13}
}

@article{zhang2024tinychart,
  title={Tinychart: Efficient chart understanding with visual token merging and program-of-thoughts learning},
  author={Zhang, Liang and Hu, Anwen and Xu, Haiyang and Yan, Ming and Xu, Yichen and Jin, Qin and Zhang, Ji and Huang, Fei},
  journal={arXiv preprint arXiv:2404.16635},
  year={2024}
}

@article{xia2024chartx,
  title={Chartx \& chartvlm: A versatile benchmark and foundation model for complicated chart reasoning},
  author={Xia, Renqiu and Zhang, Bo and Ye, Hancheng and Yan, Xiangchao and Liu, Qi and Zhou, Hongbin and Chen, Zijun and Dou, Min and Shi, Botian and Yan, Junchi and others},
  journal={arXiv preprint arXiv:2402.12185},
  year={2024}
}

@article{shi2024chartmimic,
  title={ChartMimic: Evaluating LMM's Cross-Modal Reasoning Capability via Chart-to-Code Generation},
  author={Shi, Chufan and Yang, Cheng and Liu, Yaxin and Shui, Bo and Wang, Junjie and Jing, Mohan and Xu, Linran and Zhu, Xinyu and Li, Siheng and Zhang, Yuxiang and others},
  journal={arXiv preprint arXiv:2406.09961},
  year={2024}
}

@article{zheng2023codegeex,
  title={Codegeex: A pre-trained model for code generation with multilingual evaluations on humaneval-x},
  author={Zheng, Qinkai and Xia, Xiao and Zou, Xu and Dong, Yuxiao and Wang, Shan and Xue, Yufei and Wang, Zihan and Shen, Lei and Wang, Andi and Li, Yang and others},
  journal={arXiv preprint arXiv:2303.17568},
  year={2023}
}

@inproceedings{wei2024magicoder,
  title={Magicoder: Empowering code generation with oss-instruct},
  author={Wei, Yuxiang and Wang, Zhe and Liu, Jiawei and Ding, Yifeng and Zhang, Lingming},
  booktitle={Forty-first International Conference on Machine Learning},
  year={2024}
}

@article{zhang2024humaneval,
  title={HumanEval-V: Evaluating Visual Understanding and Reasoning Abilities of Large Multimodal Models Through Coding Tasks},
  author={Zhang, Fengji and Wu, Linquan and Bai, Huiyu and Lin, Guancheng and Li, Xiao and Yu, Xiao and Wang, Yue and Chen, Bei and Keung, Jacky},
  journal={arXiv preprint arXiv:2410.12381},
  year={2024}
}

@inproceedings{Masry2024ChartInstructIT,
  title={ChartInstruct: Instruction Tuning for Chart Comprehension and Reasoning},
  author={Ahmed Masry and Mehrad Shahmohammadi and Md. Rizwan Parvez and Enamul Hoque and Shafiq R. Joty},
  booktitle={Annual Meeting of the Association for Computational Linguistics},
  year={2024},

}

@inproceedings{li2024mmcode,
  title={MMCode: Benchmarking Multimodal Large Language Models for Code Generation with Visually Rich Programming Problems},
  author={Li, Kaixin and Tian, Yuchen and Hu, Qisheng and Luo, Ziyang and Huang, Zhiyong and Ma, Jing},
  booktitle={Findings of the Association for Computational Linguistics: EMNLP 2024},
  pages={736--783},
  year={2024}
}

@article{wu2024plot2code,
  title={Plot2Code: A Comprehensive Benchmark for Evaluating Multi-modal Large Language Models in Code Generation from Scientific Plots},
  author={Wu, Chengyue and Ge, Yixiao and Guo, Qiushan and Wang, Jiahao and Liang, Zhixuan and Lu, Zeyu and Shan, Ying and Luo, Ping},
  journal={arXiv preprint arXiv:2405.07990},
  year={2024}
}

@inproceedings{zhang2024gpt,
  title={Is gpt-4v (ision) all you need for automating academic data visualization? exploring vision-language models’ capability in reproducing academic charts},
  author={Zhang, Zhehao and Ma, Weicheng and Vosoughi, Soroush},
  booktitle={Findings of the Association for Computational Linguistics: EMNLP 2024},
  pages={8271--8288},
  year={2024}
}

@inproceedings{yan2024chartreformer,
  title={Chartreformer: Natural language-driven chart image editing},
  author={Yan, Pengyu and Bhosale, Mahesh and Lal, Jay and Adhikari, Bikhyat and Doermann, David},
  booktitle={International Conference on Document Analysis and Recognition},
  pages={453--469},
  year={2024},
  organization={Springer}
}

@ARTICLE{10884879,
  author={Huang, Yi and Huang, Jiancheng and Liu, Yifan and Yan, Mingfu and Lv, Jiaxi and Liu, Jianzhuang and Xiong, Wei and Zhang, He and Cao, Liangliang and Chen, Shifeng},
  journal={IEEE Transactions on Pattern Analysis and Machine Intelligence}, 
  title={Diffusion Model-Based Image Editing: A Survey}, 
  year={2025},
  volume={47},
  number={6},
  pages={4409-4437},
  doi={10.1109/TPAMI.2025.3541625}}

@InProceedings{Brack_2024_CVPR,
    author    = {Brack, Manuel and Friedrich, Felix and Kornmeier, Katharia and Tsaban, Linoy and Schramowski, Patrick and Kersting, Kristian and Passos, Apolinario},
    title     = {LEDITS++: Limitless Image Editing using Text-to-Image Models},
    booktitle = {Proceedings of the IEEE/CVF Conference on Computer Vision and Pattern Recognition},
    month     = {June},
    year      = {2024},
    pages     = {8861-8870}
}

@inproceedings{sheynin2024emu,
  title={Emu edit: Precise image editing via recognition and generation tasks},
  author={Sheynin, Shelly and Polyak, Adam and Singer, Uriel and Kirstain, Yuval and Zohar, Amit and Ashual, Oron and Parikh, Devi and Taigman, Yaniv},
  booktitle={Proceedings of the IEEE/CVF Conference on Computer Vision and Pattern Recognition},
  pages={8871--8879},
  year={2024}
}

@inproceedings{yu2025anyedit,
  title={Anyedit: Mastering unified high-quality image editing for any idea},
  author={Yu, Qifan and Chow, Wei and Yue, Zhongqi and Pan, Kaihang and Wu, Yang and Wan, Xiaoyang and Li, Juncheng and Tang, Siliang and Zhang, Hanwang and Zhuang, Yueting},
  booktitle={Proceedings of the Computer Vision and Pattern Recognition Conference},
  pages={26125--26135},
  year={2025}
}

@inproceedings{huang2024smartedit,
  title={Smartedit: Exploring complex instruction-based image editing with multimodal large language models},
  author={Huang, Yuzhou and Xie, Liangbin and Wang, Xintao and Yuan, Ziyang and Cun, Xiaodong and Ge, Yixiao and Zhou, Jiantao and Dong, Chao and Huang, Rui and Zhang, Ruimao and others},
  booktitle={Proceedings of the IEEE/CVF Conference on Computer Vision and Pattern Recognition},
  pages={8362--8371},
  year={2024}
}

@inproceedings{liu2024towards,
  title={Towards understanding cross and self-attention in stable diffusion for text-guided image editing},
  author={Liu, Bingyan and Wang, Chengyu and Cao, Tingfeng and Jia, Kui and Huang, Jun},
  booktitle={Proceedings of the IEEE/CVF conference on computer vision and pattern recognition},
  pages={7817--7826},
  year={2024}
}

@inproceedings{han2024proxedit,
  title={Proxedit: Improving tuning-free real image editing with proximal guidance},
  author={Han, Ligong and Wen, Song and Chen, Qi and Zhang, Zhixing and Song, Kunpeng and Ren, Mengwei and Gao, Ruijiang and Stathopoulos, Anastasis and He, Xiaoxiao and Chen, Yuxiao and others},
  booktitle={Proceedings of the IEEE/CVF Winter Conference on Applications of Computer Vision},
  pages={4291--4301},
  year={2024}
}

@inproceedings{shi2024dragdiffusion,
  title={Dragdiffusion: Harnessing diffusion models for interactive point-based image editing},
  author={Shi, Yujun and Xue, Chuhui and Liew, Jun Hao and Pan, Jiachun and Yan, Hanshu and Zhang, Wenqing and Tan, Vincent YF and Bai, Song},
  booktitle={Proceedings of the IEEE/CVF Conference on Computer Vision and Pattern Recognition},
  pages={8839--8849},
  year={2024}
}

@article{wang2004ssim,
  title={Image quality assessment: From error visibility to structural similarity},
  author={Wang, Zhou and Bovik, Alan C. and Sheikh, Hamid R. and Simoncelli, Eero P.},
  journal={IEEE Transactions on Image Processing},
  volume={13},
  number={4},
  pages={600--612},
  year={2004},
  publisher={IEEE}
}

@inproceedings{hore2010psnr,
  title={Image quality metrics: PSNR vs. SSIM},
  author={Hore, Aljoscha and Ziou, Djemel},
  booktitle={20th International Conference on Pattern Recognition},
  pages={2366--2369},
  year={2010},
  organization={IEEE}
}

@inproceedings{zhang2018lpips,
  title={The unreasonable effectiveness of deep features as a perceptual metric},
  author={Zhang, Richard and Isola, Phillip and Efros, Alexei A. and Shechtman, Eli and Wang, Oliver},
  booktitle={IEEE Conference on Computer Vision and Pattern Recognition (CVPR)},
  pages={586--595},
  year={2018}
}

@inproceedings{smith2007tesseract,
  title={An overview of the Tesseract OCR engine},
  author={Smith, Ray},
  booktitle={Ninth International Conference on Document Analysis and Recognition },
  volume={2},
  pages={629--633},
  year={2007},
  organization={IEEE}
}

@article{wu2025omnigen2,
  title={OmniGen2: Exploration to Advanced Multimodal Generation},
  author={Chenyuan Wu and Pengfei Zheng and Ruiran Yan and Shitao Xiao and Xin Luo and Yueze Wang and Wanli Li and Xiyan Jiang and Yexin Liu and Junjie Zhou and Ze Liu and Ziyi Xia and Chaofan Li and Haoge Deng and Jiahao Wang and Kun Luo and Bo Zhang and Defu Lian and Xinlong Wang and Zhongyuan Wang and Tiejun Huang and Zheng Liu},
  journal={arXiv preprint arXiv:2506.18871},
  year={2025}
}

@INPROCEEDINGS{brooksinst,
  author={Brooks, Tim and Holynski, Aleksander and Efros, Alexei A.},
  booktitle={2023 IEEE/CVF Conference on Computer Vision and Pattern Recognition}, 
  title={InstructPix2Pix: Learning to Follow Image Editing Instructions}, 
  year={2023},
  volume={},
  number={},
  pages={18392-18402},

}

@misc{OpenAI_GPTImage,
  title        = {Introducing our latest image generation model in the API},
  author       = {OpenAI},
  year         = {2025},
  url = {https://openai.com/index/image-generation-api/}
}

@misc{GoogleImagen4_2025,
  title        = {Imagen 4},
  author       = {Google},
  year         = {2025},
  url = {https://deepmind.google/models/imagen/}
}

@inproceedings{li2024panoptic,
  title={Panoptic scene graph generation with semantics-prototype learning},
  author={Li, Li and Ji, Wei and Wu, Yiming and Li, Mengze and Qin, You and Wei, Lina and Zimmermann, Roger},
  booktitle={Proceedings of the AAAI conference on artificial intelligence},
  volume={38},
  number={4},
  pages={3145--3153},
  year={2024}
}

@inproceedings{li2025dpu,
  title={Dpu: Dynamic prototype updating for multimodal out-of-distribution detection},
  author={Li, Shawn and Gong, Huixian and Dong, Hao and Yang, Tiankai and Tu, Zhengzhong and Zhao, Yue},
  booktitle={Proceedings of the Computer Vision and Pattern Recognition Conference},
  pages={10193--10202},
  year={2025}
}

@article{li2025treble,
  title={Treble counterfactual vlms: A causal approach to hallucination},
  author={Li, Shawn and Qu, Jiashu and Zhou, Yuxiao and Qin, Yuehan and Yang, Tiankai and Zhao, Yue},
  journal={arXiv preprint arXiv:2503.06169},
  year={2025}
}

@article{li2025secure,
  title={Secure on-device video ood detection without backpropagation},
  author={Li, Shawn and Cai, Peilin and Zhou, Yuxiao and Ni, Zhiyu and Liang, Renjie and Qin, You and Nian, Yi and Tu, Zhengzhong and Hu, Xiyang and Zhao, Yue},
  journal={arXiv preprint arXiv:2503.06166},
  year={2025}
}

@inproceedings{limm,
author = {Li, Li and Wang, Chenwei and Qin, You and Ji, Wei and Liang, Renjie},
title = {Biased-Predicate Annotation Identification via Unbiased Visual Predicate Representation},
year = {2023},
isbn = {9798400701085},
booktitle = {Proceedings of the 31st ACM International Conference on Multimedia},
pages = {4410–4420},
numpages = {11},
}

@article{dong1,
author = {Dong, Hao and Chen, Xieyuanli and S\"{a}rkk\"{a}, Simo and Stachniss, Cyrill},
title = {Online pole segmentation on range images for long-term LiDAR localization in urban environments},
year = {2023},
issue_date = {Jan 2023},
volume = {159},
number = {C},
issn = {0921-8890},
journal = {Robot. Auton. Syst.},
month = jan,
numpages = {9},
}

@inproceedings{dong2,
author = {Dong, Hao and Nejjar, Ismail and Sun, Han and Chatzi, Eleni and Fink, Olga},
title = {SimMMDG: a simple and effective framework for multi-modal domain generalization},
year = {2023},
booktitle = {Proceedings of the 37th International Conference on Neural Information Processing Systems},
articleno = {3441},
numpages = {22},
}

@INPROCEEDINGS{liicassp,
  author={Li, Li and Qin, You and Ji, Wei and Zhou, Yuxiao and Zimmermann, Roger},
  booktitle={ICASSP 2024 - 2024 IEEE International Conference on Acoustics, Speech and Signal Processing (ICASSP)}, 
  title={Domain-Wise Invariant Learning for Panoptic Scene Graph Generation}, 
  year={2024},
  volume={},
  number={},
  pages={3165-3169},
}

@misc{li2025personalizedconversationalbenchmarksimulating,
      title={A Personalized Conversational Benchmark: Towards Simulating Personalized Conversations}, 
      author={Li Li and Peilin Cai and Ryan A. Rossi and Franck Dernoncourt and Branislav Kveton and Junda Wu and Tong Yu and Linxin Song and Tiankai Yang and Yuehan Qin and Nesreen K. Ahmed and Samyadeep Basu and Subhojyoti Mukherjee and Ruiyi Zhang and Zhengmian Hu and Bo Ni and Yuxiao Zhou and Zichao Wang and Yue Huang and Yu Wang and Xiangliang Zhang and Philip S. Yu and Xiyang Hu and Yue Zhao},
      year={2025},
      eprint={2505.14106},
      archivePrefix={arXiv},
}

@article{litomm,
author = {Ji, Wei and Li, Li and Fei, Hao and Liu, Xiangyan and Yang, Xun and Li, Juncheng and Zimmermann, Roger},
title = {Toward Complex-query Referring Image Segmentation: A Novel Benchmark},
year = {2024},
issue_date = {January 2025},
volume = {21},
number = {1},
issn = {1551-6857},
journal = {ACM Trans. Multimedia Comput. Commun. Appl.},
month = dec,
articleno = {40},
numpages = {18}
}

@inproceedings{NEURIPS2023_407106f4,
 author = {Zhang, Ao and Fei, Hao and Yao, Yuan and Ji, Wei and Li, Li and Liu, Zhiyuan and Chua, Tat-Seng},
 booktitle = {Advances in Neural Information Processing Systems},
 pages = {20299--20319},
 publisher = {Curran Associates, Inc.},
 title = {VPGTrans: Transfer Visual Prompt Generator across LLMs},
 volume = {36},
 year = {2023}
}

@inproceedings{liicse,
author = {Sun, Zhensu and Li, Li and Liu, Yan and Du, Xiaoning and Li, Li},
title = {On the importance of building high-quality training datasets for neural code search},
year = {2022},
isbn = {9781450392211},
booktitle = {Proceedings of the 44th International Conference on Software Engineering},
pages = {1609–1620},
numpages = {12}
}

@article{qin2024metaood,
  title={Metaood: Automatic selection of ood detection models},
  author={Qin, Yuehan and Zhang, Yichi and Nian, Yi and Ding, Xueying and Zhao, Yue},
  journal={arXiv preprint arXiv:2410.03074},
  year={2024}
}

@article{liu2025principled,
  title={Principled multimodal representation learning},
  author={Liu, Xiaohao and Xia, Xiaobo and Ng, See-Kiong and Chua, Tat-Seng},
  journal={arXiv preprint arXiv:2507.17343},
  year={2025}
}

@article{liu2025continual,
  title={Continual multimodal contrastive learning},
  author={Liu, Xiaohao and Xia, Xiaobo and Ng, See-Kiong and Chua, Tat-Seng},
  journal={arXiv preprint arXiv:2503.14963},
  year={2025}
}
\bibliographystyle{iclr2023_conference}

\clearpage
\appendix

\section{Extended Related Work}
\label{appendix:related}

\noindent \textbf{Text-to-Image Generation.}  
The rapid progress of diffusion-based models has revolutionized text-conditioned image generation, enabling results that are both high-fidelity and prompt-faithful \citep{ramesh2022hierarchicaltextconditionalimagegeneration,dong1, dong2, NEURIPS2023_407106f4,rombach2022highresolutionimagesynthesislatent}. ControlNet and related approaches expand controllability by incorporating structural or spatial priors \citep{zhang2023addingconditionalcontroltexttoimage}. Yet these advances have focused primarily on natural imagery. Scientific figures remain relatively neglected, despite their demand for symbolic precision, calibrated spatial relationships, and embedded textual fidelity. Evaluations show mainstream systems often fail in data accuracy and layout coherence for scientific use cases \citep{zhang2024scimagegoodmultimodallarge}. In response, specialized methods such as OCR-aware generative frameworks \citep{rodriguez2023ocr} and programmatic vector-graphic synthesis \citep{belouadi2024automatikztextguidedsynthesisscientific} highlight the need for tailored solutions.  

\noindent \textbf{Image Editing.}  
Instruction-based editing has evolved from GANs and encoder-based systems toward diffusion-driven methods, which better balance realism with semantic alignment. A survey by \citet{10884879} provides a comprehensive overview of this transition. Representative works include LEDITS++ \citep{Brack_2024_CVPR}, which extends text-driven editing to unconstrained transformations; Emu Edit \citep{sheynin2024emu}, which integrates recognition for localized precision; and Liu et al.~\citep{liu2024towards}, which probe attention mechanisms to preserve semantic fidelity. More recent works push toward interactivity and compositionality: SmartEdit \citep{huang2024smartedit} employs multimodal LLMs to compose edits, ProxEdit \citep{han2024proxedit} stabilizes transformations without tuning, and DragDiffusion \citep{shi2024dragdiffusion} enables point-based manipulation. AnyEdit \citep{yu2025anyedit} exemplifies the broader trajectory toward unified, general-purpose editing frameworks.  

\noindent \textbf{Scientific Chart Editing.}  
Unlike natural images, charts encode structured data, calibrated axes, and embedded text, requiring semantic consistency and readability throughout editing. While a broad literature addresses diffusion-based editing of natural scenes \citep{brooks2023instructpix2pixlearningfollowimage,huang2024smartedit,han2024proxedit,liicse,sheynin2024emu,limm,liicassp,shi2024dragdiffusion,yu2025anyedit,10884879}, research specific to scientific figures is limited. ScImage investigates the limitations of multimodal LLMs for figure generation \citep{zhang2024scimagegoodmultimodallarge}; AutomaTikZ explores text-to-vector generation under programmatic constraints \citep{belouadi2024automatikztextguidedsynthesisscientific}; and ChartEdit formulates chart editing as a multimodal evaluation benchmark \citep{zhao2025chartedit}. A common limitation in existing work is reliance on intermediate code (e.g., matplotlib) as the target of modification. While this guarantees structural validity, it reduces evaluation to code executability and neglects perceptual quality and user-facing usability. Thus, the field lacks benchmarks that jointly measure instruction adherence, semantic fidelity, and visual clarity in an end-to-end setting, motivating figure editing as a distinct line of inquiry.  

\section{Base Figure Sourcing and Generation }
\label{appx:basefig}

As discussed in Sec.~\ref{sec:basefig}, base figures are generated for chart classes $\mathcal{C}$ (bar, stacked–bar, line, area, box, violin, donut, pie, dot, scatter) using dataset names from a curated whitelist $\mathcal{A}$ (see Appx.~\ref{appendix:datasets}). 
For each class $c \in \mathcal{C}$, a preference list $\mathcal{P}(c)\subseteq\mathcal{A}$ guides the assignment toward semantically coherent themes. 

\paragraph{LLM–guided spec proposal.}
A chat model $M$ is instructed to output a single JSON object
\[
o=\{\texttt{vega\_spec}=\sigma,\ \texttt{dataset}=d\},\qquad \sigma\in\Sigma,\ d\in\mathcal{A},
\]
where $\mathcal{A}$ is the set of allowed dataset names. Any mismatch with the requested dataset $d$ triggers rejection and re-sampling. Each prompt includes a class hint $H(c)$, a preferred dataset list $\mathcal{P}(c)$, an exemplar specification $E_c$ (style only), and an \emph{avoid–terms} block derived from recent generations. The sampling temperature is fixed to $\tau=0.55$ to balance validity and diversity. Detailed prompt template is shown below:

\begin{tcolorbox}[colback=gray!5!white,colframe=black!50,title={System Message (Part 1)}]
\label{appx:gen-prompts}
\begin{lstlisting}
You return ONLY a valid JSON object with exactly two fields:
  - vega_spec: a VALID Vega v6 specification JSON object
  - dataset: ONE string chosen ONLY from the allowed set (see Appendix A, Table X)

Return exactly one JSON object. Do not add any prefix, suffix, or code fences.
No Markdown, no explanations, no backticks.

vega_spec requirements:
  - Use "$schema": "https://vega.github.io/schema/vega/v6.json".
  - Base data on a popular public dataset; URLs are disallowed. Include a small
    inline sample in "values".
  - The chosen dataset MUST naturally support category->value or
    category x series->value aggregation suitable for bar/stacked/grouped charts.
  - Match the requested chart class:
      * bar: one categorical field + one quantitative field
      * stacked-bar: category + series + value (stacked)
      * grouped-bar: category + series + value (side-by-side)
  - Include all necessary components (data, scales, axes, marks) so it renders.
  - Do NOT include extra meta fields inside vega_spec.
\end{lstlisting}
\end{tcolorbox}

\begin{tcolorbox}[colback=gray!5!white,colframe=black!50,title={System Message (Part 2)}]
\begin{lstlisting}
Self-check before responding:
  - Prefer a specific dataset name from the allowed set that credibly aligns
    with field/entity names (full whitelist in Appendix A, Table X, see \ref{appx:allowed-datasets}).
  - Use "unknown" only if no credible alignment exists.

Output rules:
  - Return EXACTLY one JSON object containing { "vega_spec": ..., "dataset": ... }.
  - vega_spec must be a valid Vega v6 JSON object with inline "values" (no "url").
  - No Markdown, no commentary, no extra keys.
\end{lstlisting}
\end{tcolorbox}

\begin{tcolorbox}[colback=gray!5!white,colframe=black!50,title={User Template (Bar, excerpt)}]
\begin{lstlisting}
Task:
- Chart class: {chart_class}
- Instruction: {class_hint}
- Generate a JSON object with two fields: vega_spec (Vega v6 spec) and dataset
  (string from the allowed set or 'unknown').

Hard constraint for this sample:
- You MUST set "dataset" EXACTLY to: {dataset_target}
- Field names and inline 'values' must be plausible for {dataset_target}.

Data requirements:
- Include inline 'values' only (no 'url'); fit bar => category + value.
- Use 5..12 categories; numeric magnitudes should be plausible for the dataset.

Diversity controls:
- Avoid reusing identical numeric multisets for the same field set.
- Avoid terms seen recently:
{avoid_terms}

Preferred datasets for this class (see Appendix A, Table X for the full list):
{preferred_datasets}

Output:
- ONLY one JSON object: { "vega_spec": ..., "dataset": ... }.
\end{lstlisting}
\end{tcolorbox}

\paragraph{Scheduling and validity.}
A scheduler balances dataset usage by always selecting the least-used candidate for each chart class, based on compatibility heuristics (e.g., time series $\mapsto$ line/area; survey data $\mapsto$ bar/pie/dot). Returned specifications are checked for Vega v6 schema, completeness (\texttt{data}, \texttt{marks}, \texttt{scales}, \texttt{axes}), and type-specific field patterns (e.g., bar requires \{category, numeric\}; stacked–bar requires \{category, series, numeric\}). Invalid proposals are rejected and resampled.

\paragraph{Shape validation.}
Beyond generic schema checks, additional constraints enforce meaningful content. For example, bar charts must contain at least one categorical and one numeric field, while stacked–bar charts must include two categorical fields and one numeric field. Other chart types are validated using generic rules.

\paragraph{Duplicate and near–duplicate control.}
For every $\sigma$, we compute four signatures over its inline data and structure:
\[
\begin{aligned}
h_{\mathrm{exact}}(\sigma) &= \mathrm{SHA256}\!\left(\bigoplus\nolimits_{\text{rows}}\ \text{sorted Vega rows}\right),\\
h_{\mathrm{multi}}(\sigma) &= \mathrm{SHA256}\!\left(\text{numeric multiset per field set}\right),\\
s_{\mathrm{val}}(\sigma) &= \mathrm{SHA256}\!\left(\text{per–field histograms with }b{=}6\text{ and }(\mu,\sigma)\right),\\
s_{\mathrm{struct}}(\sigma) &= \mathrm{SHA256}\!\left(\text{size buckets, mark types, scale types/flags, axis orients, legend presence}\right),
\end{aligned}
\]
where SHA256 is a cryptographic hash function that produces a fixed 256-bit digest with extremely low collision probability. 
A specification is rejected if $h_{\mathrm{exact}}$ or $h_{\mathrm{multi}}$ has been observed previously, or if both $s_{\mathrm{val}}$ and $s_{\mathrm{struct}}$ have appeared before. 
This eliminates duplicates and near–duplicates while permitting controlled variability.

\paragraph{Semantic diversity via term overlap.}
Categorical fields are inferred from scales and encodings, forming a token set $T(\sigma)$. 
A sliding window $\mathsf{W}$ of the last $k$ samples (default $k=16$) is maintained, and the Jaccard overlap ratio
\[
r=\frac{|T(\sigma)\cap U|}{|T(\sigma)\cup U|},\qquad U=\bigcup_{S\in\mathsf{W}}S
\]
is computed. 
A candidate is rejected if $r>\theta$ (default $\theta=0.70$) and $|T(\sigma)\setminus U|< m$ (default $m=2$). 
The current union $U$ is injected back into subsequent prompts as an \emph{avoid–terms} block, enforcing semantic diversity across generations.

\paragraph{Additional mechanisms.}
Further enhancements improve robustness: malformed completions are handled by stripping Markdown fences or extracting JSON blocks; provenance is logged into a JSON index with raw outputs for debugging. 
Together, these mechanisms ensure quality, diversity, and reproducibility of the generated base figures.

\section{Editing Operations}
\label{sec:ops_appendix}

As briefly discussed in Sec.~\ref{sec:ops}, this section provides extended details of the editing operations and annotation pipeline. 
We describe how we generated single and multi-edit supervision, conversational annotations, visual-guidance assets, and style-transfer pairs. 
Representative prompt excerpts are also included.
We first define a canonical operation set:
\[
\mathcal{O}=\left\{
\begin{array}{ll}
\texttt{change\_datapoint\_color}, & \texttt{increase\_text\_size}, \\
\texttt{decrease\_text\_size}, & \texttt{change\_background\_color}, \\
\texttt{increase\_category\_spacing}, & \texttt{decrease\_category\_spacing}, \\
\texttt{add\_title}, & \texttt{add\_datapoint}, \\
\texttt{remove\_datapoint}
\end{array}
\right\}.
\]

\subsection{Single \& Multi Edit Generation}
\label{sec:ops:gen}
For each chart specification we select a feasible subset $\mathcal{O}(c)$, filtering out inapplicable edits 
(e.g., spacing operations for charts without band/point scales, or removals when only one data row exists). 
An LLM is prompted to return exactly one sentence instruction followed by explicit OP tags, as well as the edited Vega~v6 specification. 
We canonicalize op names, infer missing keys (such as \texttt{axis\_label\_size}, \texttt{new\_color}, \texttt{new\_bg}, or \texttt{new\_padding}), and apply minimal but deterministic edits to ensure the modification is visually effective. 
Validation includes schema conformance, key completeness, and visible effect realization. Detailed prompt is shown below:

\begin{tcolorbox}[colback=gray!5!white,colframe=black!50,title={Prompt (Single/Multi, Part 1)}]
\begin{lstlisting}
Return ONLY a JSON object (no markdown) with keys:
- "instruction": ONE English sentence that describes exactly N edits (N in {1,2,3}),
  followed by EXACTLY N tag lines in order:
    [#OP1 op=<...>; key1=value1; key2=value2; ...]
    [#OP2 op=<...>; ...]
    [#OP3 op=<...>; ...]  (only if N==3)
- "ops": array of length N; each item has "op" in:
  ["change_datapoint_color","increase_text_size","decrease_text_size",
   "change_background_color","increase_category_spacing","decrease_category_spacing",
   "add_title","add_datapoint","remove_datapoint"]
- "edited_spec": a VALID Vega v6 JSON spec that keeps "$schema" v6 and uses inline "values" only
  (strictly no "url").

Required keys per operation (inside each [#OPi ...] tag):
- change_background_color:
    new_bg=<css-or-#hex>
- increase_text_size / decrease_text_size:
    axis_label_size=<int in 6..30>; tick_size=<int> (optional); title_size=<int> (optional)
\end{lstlisting}
\end{tcolorbox}

\begin{tcolorbox}[colback=gray!5!white,colframe=black!50,title={Prompt (Single/Multi, Part 2)}]
\begin{lstlisting}
- increase_category_spacing / decrease_category_spacing:
    scale=auto|x|y; new_padding=<float in [0,0.9]>
- add_title:
    title_text=<non-empty>; title_size=<int> (optional)
- change_datapoint_color:
    target_category=<label>; target_series=<label> (omit if not applicable);
    new_color=<css-or-#hex not used in original>
- add_datapoint:
    If single-series:
      position=before:<existing>|after:<existing>|end;
      category=<new_label>; value=<number>
    If multi-series:
      position=...; category=<cat_label>; series=<series_label>; value=<number>
- remove_datapoint:
    If single-series:
      category=<existing_label>
    If multi-series:
      category=<existing_label>; series=<existing_series>

Editing rules and checks:
- Apply minimal, deterministic edits; ensure each step has a visible effect.
- Maintain inline data only; never introduce "url".
- Sizes must be >= 6; band/point padding must be within [0,0.9].
- New colors must not collide with original literal colors.
- For add/remove datapoint, edit exactly one row and keep row-count accounting consistent:
    rows(edited) = rows(original) + adds - removes
- Preserve unrelated content (data, labels, titles) unless the step explicitly changes them.

For Single set N=1; for Multi set N in {2,3}. Output exactly one JSON object.
\end{lstlisting}
\end{tcolorbox}

\subsection{Conversational Annotations}
\label{sec:ops:combo2}
To simulate multi-turn editing, we align each two-op edit with its corresponding single-op edits. 
Given a two-op edit $(o_1,o_2)$, we locate the two single edits with the same operations, generate intermediate ground truth images, and concatenate them into a two-round dialogue. 
Each conversational sample therefore contains: (i) the original figure, (ii) turn-1 with an instruction and intermediate ground truth, and (iii) turn-2 with a follow-up instruction and the final ground truth. 
This design yields per-round supervision and enables evaluation of temporal consistency. Detailed prompt is shown below:

\begin{tcolorbox}[colback=gray!5!white,colframe=black!50,title={Prompt (Conversation)}]
\begin{lstlisting}
You are a strict formatter that assembles a 2-turn conversation object
from provided edit annotations. You MUST return ONLY one JSON object.

Inputs (conceptual):
- original: the unedited chart (spec+image).
- single-edits: two entries, each has:
    op (atomic op name),
    instruction (one sentence + [#OP* ...] tags),
    edited_spec (Vega v6, inline values only),
    edited_image.
- multi-edit (2-step): has:
    ops=[op1, op2] in the exact execution order,
    edited_spec (final target),
    edited_image,
    instruction (one sentence + tags).

Formatting rules:
1) Preserve execution order strictly as [op1, op2].
2) Use the single-edits whose op matches op1 and op2, respectively.
3) For each turn i in {0,1}:
   - instruction: copy the corresponding single's instruction verbatim,
     trimming leading/trailing whitespace only.
   - gt: spec=image=the corresponding single's ground truth (intermediate).
   - op: the corresponding op (op1 for turn 0, op2 for turn 1).
4) final: use the multi-edit's edited_spec and edited_image.
5) Include the multi-edit's instruction as multi_instruction.
6) Do NOT invent or rewrite text; do NOT change specs.
7) Output must be a single JSON object with the following fields:

{
  "chart_type": "<string>",
  "figure_id": "<string>",
  "ops": ["<op1>", "<op2>"],
  "original": {"spec": <json>, "image": "<path-or-id>"},
  "turns": [
    {
      "turn_idx": 0,
      "op": "<op1>",
      "instruction": "<single1_instruction_trimmed>",
      "gt": {"spec": <json>, "image": "<path-or-id>"}
    },
    {
      "turn_idx": 1,
      "op": "<op2>",
      "instruction": "<single2_instruction_trimmed>",
      "gt": {"spec": <json>, "image": "<path-or-id>"}
    }
  ],
  "final": {"spec": <json>, "image": "<path-or-id>"},
  "multi_instruction": "<multi_instruction_trimmed>"
}

Return exactly this one JSON object and nothing else.
\end{lstlisting}
\end{tcolorbox}

\subsection{Visual–Guidance Assets}
\label{sec:ops:visual}
For selected atomic operations (notably datapoint color changes and datapoint removals), we construct visual-guided variants by highlighting the target region directly in the original chart. 
To produce the overlays, we employ a vision–language model (GPT-Image) instructed to draw a thin red circle around the specified element while leaving the rest of the chart untouched. 
This yields paired data: (i) a natural-language instruction referencing the circled element, and (ii) a visually annotated chart. 
Such assets allow evaluation of multimodal understanding, where the model must integrate textual instructions with explicit visual cues. Detailed prompt is shown below:

\begin{tcolorbox}[colback=gray!5!white,colframe=black!50,title={Prompt (Visual Guidance)}]
\begin{lstlisting}
You are an image editor. Given a chart image and a target description,
draw a thin red circle around exactly one target element. Do not change
any chart content.

Inputs:
- Image: the original chart may be letterboxed on a plain background.
- Chart noun: {bar|slice|point|mark}.
- Target condition (optional but preferred):
    category == "<CATEGORY>"
    series   == "<SERIES>"

Task:
- Locate the single element that satisfies the target condition (if given).
- If no explicit condition is given, use the instruction sentence prefix
  as a hint to identify the most likely target element.
- Draw exactly one circle that tightly encloses the target element.

Rendering constraints:
- Stroke color: #FF0000 (pure red).
- Stroke style: thin line, no glow, no shadow.
- Circle only; no arrows, no text, no highlights or masks.
- Do not crop, scale, or move the chart content.
- If the image is letterboxed, ignore padding/margins/borders and place
  the circle over the chart area only.
- Preserve the original resolution and aspect ratio.
- Do not alter colors, fonts, or data marks other than the added circle.

Output:
- Return a single edited image where the only modification is the thin
  red circle tightly around the target element.
\end{lstlisting}
\end{tcolorbox}

\subsection{Style–Transfer Singles}
\label{sec:ops:transfer}
We further derive one-shot style-transfer supervision by linking existing single edits to style sources. 
For each single edit, we identify another original chart whose current style attribute already matches the target attribute of the edited chart. 
We construct a natural instruction such as “Make this bar chart use the same background color as the reference chart,” and pair it with the corresponding OP tag. 
This produces style-transfer pairs across both same-type and cross-type chart classes, enabling evaluation of style generalization. Detailed prompt is shown below:

\begin{tcolorbox}[colback=gray!5!white,colframe=black!50,title={Prompt (Style Transfer)}]
\begin{lstlisting}
You are writing ONE natural-language instruction for a style transfer (single attribute).
Inputs you are given (out of band) include:
- target chart class and image,
- a SINGLE-OP edit already chosen for the target (with its OP tag line),
- a reference chart ("the reference chart") that already exhibits the target style value.

Task:
- Write exactly ONE concise English sentence that asks to make the target chart
  use the SAME style attribute as the reference chart.
- Use phrasing like:
    "Make this <chart noun> use the same background color as the reference chart."
    "Make this <chart noun> use the same axis label font size as the reference chart."
    "Make this <chart noun> use the same category spacing as the reference chart."
    "Match the datapoint color used in the reference chart."
    "Add a title with the same font size as the reference chart."
- Do NOT mention internal ids. Say "the reference chart" or "the example chart".

After the sentence, append EXACTLY ONE OP tag line, keeping the provided tag VERBATIM:
  [#OP1 op=<one_of_allowed_ops>; key1=value1; key2=value2; ...]

Allowed ops (single attribute only):
  change_background_color | increase_text_size | decrease_text_size |
  increase_category_spacing | decrease_category_spacing |
  add_title | change_datapoint_color

Hard constraints:
- Output ONLY the final instruction text (one sentence) followed by the single OP tag line.
- Do NOT return JSON. Do NOT include edited_spec. Do NOT invent new keys or values.
- Preserve the original OP tag exactly as given (verbatim).
\end{lstlisting}
\end{tcolorbox}

Through these pipelines, each figure can appear as (i) atomic edits (single/multi), 
(ii) conversational trajectories, (iii) visually guided variants, and 
(iv) style-transfer pairs. 
All assets are designed to be reproducible, diverse, and machine-readable, 
while supporting multimodal evaluation settings.

\section{Evaluation Metrics}
\label{app:metrics}

As discussed in Sec.~\ref{sec:evaluation}, we report both classic image metrics and an LLM-based score to capture semantic correctness.

\noindent \textbf{SSIM.} Structural Similarity Index~\cite{wang2004ssim} is applied on grayscale renderings with Gaussian weighting to emphasize local structure. This metric accounts for luminance, contrast, and structure, making it more perceptually meaningful than raw pixel errors.

\noindent \textbf{PSNR.} Peak Signal–to–Noise Ratio~\cite{hore2010psnr} is computed with pixel values clipped to $[0,255]$ and averaged across RGB channels. It quantifies the logarithmic ratio between the maximum possible signal and mean squared error.

\noindent \textbf{LPIPS.} Learned Perceptual Image Patch Similarity~\cite{zhang2018lpips} is computed using the official framework, with AlexNet as the default backbone. Images are normalized to $[-1,1]$ before feature extraction. LPIPS captures perceptual discrepancies such as texture or shape distortions.

\noindent \textbf{CLIP similarity.} We use CLIP ViT-L/14~\cite{radford2021learning} to extract image embeddings and report cosine similarity between $\widehat{I}$ and $I^\star$. This provides a semantic-level measure of alignment beyond pixel similarity.

\noindent \textbf{OCR similarity.} We extract text from both images using Tesseract OCR~\cite{smith2007tesseract}. Similarity is measured as the normalized edit distance:
\[
\mathrm{Sim}_{\text{OCR}} = 1 - \frac{\mathrm{EditDist}(s_{\widehat{I}}, s_{I^\star})}{\max(|s_{\widehat{I}}|, |s_{I^\star}|)},
\]
where $s_{\widehat{I}}$ and $s_{I^\star}$ are the concatenated OCR strings. This metric emphasizes correctness of labels, legends, and annotations.

\noindent \textbf{LLM-based instruction score.} To directly evaluate editing success, we prompt a large language model~\cite{openai2024gpt4o} with (i) the original chart and instruction $(I,u)$, (ii) the edited output $\widehat{I}$, and (iii) the ground truth $I^\star$. The model issues binary judgments on:
\begin{itemize}
    \item Instruction satisfaction: whether the requested edit is applied.
    \item Content preservation: whether the underlying chart data remain intact.
    \item Visual quality: whether the rendering is artifact-free and coherent.
\end{itemize}
Responses are parsed into structured JSON objects, which are aggregated into per-instance and per-model scores. Trimmed prompt examples are provided below:

\begin{tcolorbox}[colback=gray!5!white,colframe=black!50,title={LLM Score (System Message, Part 1)}]
\begin{lstlisting}
You are an expert AI assistant specializing in data visualization evaluation.
Your task is to evaluate how well an AI-generated chart follows a given text instruction. 
You will be given an instruction, a reference "Ground Truth" image, and the "Generated Image" to evaluate.

Evaluate the generated image based on the following three criteria:

1. Instruction Following (score_instruction): How accurately was the specific 
   instruction executed? (e.g., if asked to change color to orange, is it orange?)
2. Content Preservation (score_preservation): Were all other elements of the 
   chart preserved correctly without unwanted changes? (e.g., data values, 
   labels, and titles are unchanged unless specified).
3. Image Quality (score_quality): Is the generated image free of major artifacts, 
   distortions, or unreadable text?
\end{lstlisting}
\end{tcolorbox}

\begin{tcolorbox}[colback=gray!5!white,colframe=black!50,title={LLM Score (System Message, Part 2)}]
\begin{lstlisting}
For each of the above, assign a score from 1 (very poor) to 5 (excellent).  
Then compute a total score (score) as the average of the three above, 
rounded to the nearest integer.

Your response MUST be a JSON object with the following keys:
- "score_instruction": Integer [1-5]
- "score_preservation": Integer [1-5]
- "score_quality": Integer [1-5]
- "score": Integer [1-5], the average of the above
- "reasoning": One-sentence explanation justifying the scores

Example Response:
{
    "score_instruction": 5,
    "score_preservation": 4,
    "score_quality": 5,
    "score": 5,
    "reasoning": "The instruction was followed perfectly, content was mostly 
    preserved, and the image quality is excellent."
}
\end{lstlisting}
\end{tcolorbox}

\begin{tcolorbox}[colback=gray!5!white,colframe=black!50,title={LLM Score (User Template)}]
\begin{lstlisting}
**Instruction:**
<instruction text>

**Reference Image (Ground Truth):**
<data:image/png;base64,...>

**Generated Image (to be evaluated):**
<data:image/png;base64,...>
\end{lstlisting}
\end{tcolorbox}

\section{Additional Experimental Details}
\label{appx:exp-details}
\noindent \textbf{Pre- and Post-Processing.}
To ensure consistent inputs, all charts are letterboxed into a square canvas before inference. 
After editing, outputs are mapped back to the original resolution using contain resizing, which preserves the full layout without cropping. 
This procedure guarantees that models are evaluated under identical geometric conditions while avoiding distortion of axes or labels.  

\noindent \textbf{Prompt Construction.}
For all tasks, prompts explicitly instruct the model to make localized modifications while leaving unrelated elements unchanged. 
In Visual tasks, prompts additionally emphasize that only the circled region should be modified. 
For Transfer tasks, the prompt specifies a two-panel setup, where only the left (base) panel is editable and the right (reference) panel serves as a style guide.  

\noindent \textbf{Baselines.}
\label{app:baselines}
For comparison, we include four representative baselines that capture the current state of instruction-driven image editing:
(1) GPT-Image~\citep{OpenAI_GPTImage}. 
A commercial instruction–driven editing system provided by OpenAI. 
It supports free-form natural language instructions and has been widely used for general-purpose editing tasks. 
Although proprietary, it reflects the strongest available commercial option.
(2) Imagen~4~\citep{GoogleImagen4_2025}. 
A proprietary diffusion–based editor developed by Google and released via the Vertex AI platform. 
Imagen~4 is optimized for controllable, high-fidelity image generation and editing, though its design is primarily tuned for natural image content.
(3) OmniGen~2~\citep{wu2025omnigen2}. 
An open–source multimodal model recently introduced for text-guided and image-guided editing. 
It supports multi-turn interaction and has shown promising results for chart and figure editing. 
We use the official released checkpoint and inference pipeline.
(4) InstructPix2Pix~\citep{brooksinst}. 
An open–source approach that finetunes a diffusion backbone on paired instruction–image data. 
It was among the first methods to explicitly align natural language instructions with image translation, and remains a strong research baseline for instruction-conditioned editing.

Together, these baselines span both closed and open ecosystems, diffusion and multimodal paradigms, and commercial and academic settings. 
They represent the strongest available instruction-driven editing approaches at the time of writing.

\section{More Results}
\label{appx:more_res}

\begin{figure}[t]
\centering
\includegraphics[width=\textwidth]{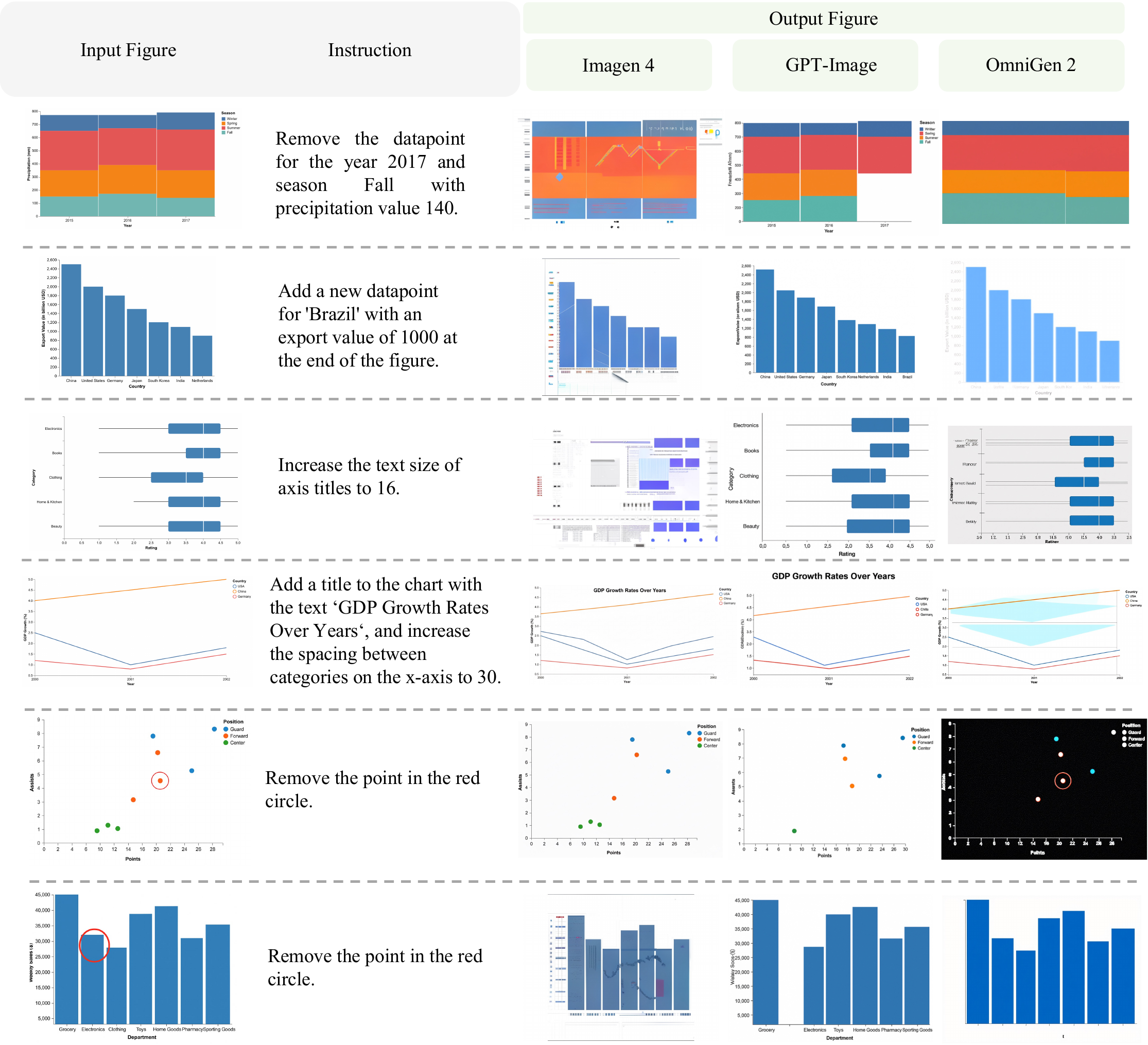} 
\vspace{-0.1in}
   \caption{Additional qualitative examples of figure editing results. Each row shows an input figure (left), the corresponding natural language instruction (middle), and the output figures generated by Imagen 4, GPT-Image, and OmniGen 2 (right). The cases cover representative edit types, including data point removal, data point addition, axis text scaling, layout adjustments, and targeted point deletion. While the models sometimes produce visually consistent outputs, they often fail to accurately execute the requested transformation, highlighting the limitations of current instruction-based figure editing systems.}
\label{fig:add_qualitative}
\end{figure}

As we discussed in Sec.~\ref{sec:mainres}, aggregate results already show a clear gap between pixel similarity and semantic correctness. 
Tab.~\ref{tab:per_instruction_metrics_part1} and Tab.~\ref{tab:per_instruction_metrics_part2}  provide a more fine-grained view, breaking down performance by specific instruction types. 

A recurring pattern is that edits involving numbers, such as adding or adjusting datapoints, are often the hardest to get right. 
Models may place a new bar or point, but the actual value is off, the axis scale shifts incorrectly, or the legend does not update. 
Edits that change the overall layout or chart type also tend to expose structural weaknesses: grouped bars converted to stacked bars often result in overlapping marks, or the scales fail to adjust. 

By contrast, stylistic edits like changing background colors are sometimes handled better, though even here models often stop short of a full update. For example, the background changes, but the legend or axis elements remain inconsistent. 
Text edits such as axis labels or titles show the partial benefit of OCR, but issues like misplaced text, font mismatches, or truncated labels still appear.

\begin{table*}[t]
\centering
\caption{Per-instruction performance comparison (Part 1/2). Higher is better for SSIM, CLIP, PSNR, OCR, and LLM Scores. Lower is better for LPIPS. \textbf{Instr.} denotes instruction following score. \textbf{Preserv.} denotes content preservation score. \textbf{Qual.} denotes image quality score.}
\label{tab:per_instruction_metrics_part1}
\footnotesize
\setlength{\tabcolsep}{4pt}
\begin{tabular}{lcccccccc}
\toprule
& \textbf{SSIM} $\uparrow$ & \textbf{LPIPS} $\downarrow$ & \textbf{CLIP} $\uparrow$ & \textbf{PSNR} $\uparrow$ & \textbf{OCR} $\uparrow$ & \multicolumn{3}{c}{\textbf{LLM Score (1--5)} $\uparrow$} \\
\cmidrule(lr){7-9}
\textbf{Model} & & & & & & \textbf{Instr.} & \textbf{Preserv.} & \textbf{Qual.} \\
\midrule
\multicolumn{9}{l}{\textbf{Instruction: Change the colors of the data point}} \\
\midrule
InstructPix2Pix   & 0.736 & 0.52 & 0.84 & 10.7 & 0.25 & 2.66 & 2.76 & 2.76 \\
OmniGen2          & 0.748 & 0.48 & 0.85 & 11.1 & 0.26 & 3.63 & 3.32 & 3.66 \\
GPT-Image         & 0.733 & 0.54 & 0.87 & 10.4 & 0.22 & 4.34 & 3.84 & 4.29 \\
Imagen 4          & 0.772 & 0.41 & 0.80 & 13.1 & 0.08 & 2.09 & 1.84 & 2.72 \\
\midrule
\multicolumn{9}{l}{\textbf{Instruction: Add a new title}} \\
\midrule
InstructPix2Pix   & 0.741 & 0.47 & 0.83 & 11.0 & 0.17 & 1.09 & 2.91 & 3.07 \\
OmniGen2          & 0.744 & 0.46 & 0.84 & 11.2 & 0.29 & 3.34 & 3.14 & 3.36 \\
GPT-Image         & 0.728 & 0.53 & 0.88 & 10.5 & 0.36 & 4.91 & 4.43 & 4.41 \\
Imagen 4          & 0.769 & 0.40 & 0.79 & 13.0 & 0.07 & 1.00 & 1.41 & 2.07 \\
\midrule
\multicolumn{9}{l}{\textbf{Instruction: Increase font size}} \\
\midrule
InstructPix2Pix   & 0.735 & 0.50 & 0.84 & 10.8 & 0.26 & 2.48 & 2.10 & 2.83 \\
OmniGen2          & 0.747 & 0.47 & 0.85 & 11.1 & 0.30 & 2.12 & 3.02 & 3.27 \\
GPT-Image         & 0.729 & 0.52 & 0.86 & 10.3 & 0.27 & 4.05 & 4.05 & 4.40 \\
Imagen 4          & 0.771 & 0.39 & 0.81 & 13.2 & 0.26 & 1.74 & 1.50 & 2.26 \\
\midrule
\multicolumn{9}{l}{\textbf{Instruction: Decrease font size}} \\
\midrule
InstructPix2Pix   & 0.748 & 0.49 & 0.85 & 11.1 & 0.27 & 2.02 & 1.77 & 2.58 \\
OmniGen2          & 0.752 & 0.46 & 0.84 & 11.4 & 0.31 & 2.10 & 3.05 & 3.38 \\
GPT-Image         & 0.734 & 0.51 & 0.86 & 10.6 & 0.24 & 2.70 & 4.15 & 4.00 \\
Imagen 4          & 0.773 & 0.38 & 0.81 & 13.2 & 0.18 & 1.61 & 1.50 & 2.18 \\
\bottomrule
\end{tabular}
\end{table*}

\begin{table*}[t]
\centering
\caption{Per-instruction performance comparison (Part 2/2). Higher is better for SSIM, CLIP, PSNR, OCR, and LLM Scores. Lower is better for LPIPS. \textbf{Instr.} denotes instruction following score. \textbf{Preserv.} denotes content preservation score. \textbf{Qual.} denotes image quality score.}
\label{tab:per_instruction_metrics_part2}
\footnotesize
\setlength{\tabcolsep}{4pt}
\begin{tabular}{lcccccccc}
\toprule
& \textbf{SSIM} $\uparrow$ & \textbf{LPIPS} $\downarrow$ & \textbf{CLIP} $\uparrow$ & \textbf{PSNR} $\uparrow$ & \textbf{OCR} $\uparrow$ & \multicolumn{3}{c}{\textbf{LLM Score (1--5)} $\uparrow$} \\
\cmidrule(lr){7-9}
\textbf{Model} & & & & & & \textbf{Instr.} & \textbf{Preserv.} & \textbf{Qual.} \\
\midrule
\multicolumn{9}{l}{\textbf{Instruction: Increase margin}} \\
\midrule
InstructPix2Pix   & 0.726 & 0.49 & 0.83 & 10.9 & 0.25 & 2.70 & 2.55 & 3.35 \\
OmniGen2          & 0.739 & 0.47 & 0.84 & 11.1 & 0.29 & 2.75 & 2.90 & 3.75 \\
GPT-Image         & 0.731 & 0.52 & 0.87 & 10.3 & 0.22 & 2.95 & 3.60 & 4.05 \\
Imagen 4          & 0.769 & 0.41 & 0.79 & 13.0 & 0.08 & 2.37 & 1.68 & 2.42 \\
\midrule
\multicolumn{9}{l}{\textbf{Instruction: Decrease margin}} \\
\midrule
InstructPix2Pix   & 0.728 & 0.48 & 0.84 & 11.2 & 0.27 & 2.60 & 2.60 & 3.55 \\
OmniGen2          & 0.742 & 0.46 & 0.83 & 11.4 & 0.30 & 2.30 & 2.80 & 3.65 \\
GPT-Image         & 0.733 & 0.51 & 0.86 & 10.5 & 0.23 & 3.15 & 3.50 & 4.05 \\
Imagen 4          & 0.771 & 0.40 & 0.80 & 13.1 & 0.09 & 2.00 & 1.56 & 2.75 \\
\midrule
\multicolumn{9}{l}{\textbf{Instruction: Add a new data point}} \\
\midrule
InstructPix2Pix   & 0.724 & 0.50 & 0.82 & 10.8 & 0.24 & 1.21 & 2.66 & 3.14 \\
OmniGen2          & 0.737 & 0.48 & 0.83 & 11.0 & 0.28 & 1.86 & 2.10 & 3.14 \\
GPT-Image         & 0.730 & 0.53 & 0.87 & 10.4 & 0.21 & 3.07 & 3.69 & 4.07 \\
Imagen 4          & 0.768 & 0.42 & 0.79 & 13.2 & 0.08 & 1.04 & 1.61 & 2.39 \\
\midrule
\multicolumn{9}{l}{\textbf{Instruction: Remove an existing data point}} \\
\midrule
InstructPix2Pix   & 0.727 & 0.49 & 0.83 & 11.1 & 0.25 & 1.59 & 1.83 & 2.83 \\
OmniGen2          & 0.740 & 0.47 & 0.82 & 11.3 & 0.27 & 1.38 & 1.59 & 2.41 \\
GPT-Image         & 0.732 & 0.52 & 0.86 & 10.5 & 0.22 & 3.10 & 3.34 & 4.28 \\
Imagen 4          & 0.770 & 0.40 & 0.80 & 13.0 & 0.07 & 1.41 & 1.68 & 2.41 \\
\bottomrule
\end{tabular}
\end{table*}

\section{Datasets Used for Base Figures}
\label{appendix:datasets}

Tab.~\ref{tab:allowed_datasets_a} and Tab.~\ref{tab:allowed_datasets_b} list all datasets from which we sampled base figures. 
These sources span public machine learning repositories, official statistical agencies, open data portals, and journalism/sports archives. 
We include the identifier strings exactly as used in our pipeline.

\begin{table}[htbp]
\centering
\caption{Allowed datasets (part A). See Table~\ref{tab:allowed_datasets_b} for continuation.}
\label{tab:allowed_datasets_a}
\begin{tabular}{p{0.95\textwidth}}
\toprule
\textbf{Datasets (Part A)} \\
\midrule
\begin{minipage}{0.95\textwidth}
\scriptsize
\begin{verbatim}
Kaggle: Titanic
Kaggle: House Prices
Kaggle: Instacart Market Basket
Kaggle: NYC Taxi Trip Duration
Kaggle: Amazon Reviews
Kaggle: Yelp Reviews
Kaggle: IMDB Reviews
Kaggle: Mercari Price Suggestion
Kaggle: Quora Insincere Questions
Kaggle: Toxic Comment Classification
Kaggle: Porto Seguro Safe Driver
Kaggle: Santander Customer Transaction
Kaggle: Santander Value Prediction
Kaggle: Global Temperature Time Series
Kaggle: COVID-19 Global Dataset
Kaggle: World Happiness Report
Kaggle: FIFA Player Statistics
Kaggle: Air Quality UCI
Kaggle: US Accidents Dataset
Kaggle: Zomato Restaurants Dataset
Kaggle: Video Game Sales
Kaggle: Netflix Movies and TV Shows
Kaggle: New York City Airbnb Open Data
Kaggle: Google Play Store Apps
Kaggle: Bike Sharing Demand
Kaggle: Rossmann Store Sales
Kaggle: Store Item Demand Forecasting Challenge
Kaggle: Walmart Recruiting - Store Sales Forecasting
Kaggle: Retailrocket Recommender System Dataset
Kaggle: 311 Service Requests - NYC
Kaggle: Chicago Crime
Kaggle: Austin Bikeshare Trips
Kaggle: Seattle Weather
Kaggle: Daily Delhi Climate
Kaggle: US Economic Indicators
Kaggle: S&P 500 Companies and Prices
Kaggle: Times Higher Education World University Rankings
Kaggle: Global Terrorism Database
Kaggle: World Development Indicators
Kaggle: Airline On-Time Performance
Kaggle: Avito Demand Prediction
Kaggle: TalkingData AdTracking Fraud Detection
Kaggle: IEEE-CIS Fraud Detection
Kaggle: Home Credit Default Risk
Kaggle: Give Me Some Credit
Kaggle: Loan Prediction III
Kaggle: Credit Card Fraud Detection
Kaggle: Telco Customer Churn
Kaggle: Bank Marketing
Kaggle: Student Performance
Kaggle: Heart Disease UCI
Kaggle: Breast Cancer Wisconsin (Diagnostic)
Kaggle: Pima Indians Diabetes Database
Kaggle: Stroke Prediction Dataset
Kaggle: FIFA 19 Player Dataset
Kaggle: NBA Player Stats
Kaggle: International Football Results
Kaggle: European Soccer Database
Kaggle: 120 years of Olympic history (athletes & results)
Kaggle: Netflix Stock Price
Kaggle: Bitcoin Historical Data
Kaggle: Cryptocurrency Historical Prices
\end{verbatim}
\end{minipage} \\
\bottomrule
\end{tabular}
\end{table}

\begin{table}[htbp]
\centering
\caption{Allowed datasets (part B). Continuation of Table~\ref{tab:allowed_datasets_a}.}
\label{tab:allowed_datasets_b}
\begin{tabular}{p{0.95\textwidth}}
\toprule
\textbf{Datasets (Part B)} \\
\midrule
\begin{minipage}{0.95\textwidth}
\scriptsize
\begin{verbatim}
UCI: Iris
UCI: Wine
UCI: Adult
UCI: Car Evaluation
UCI: Abalone
UCI: Seeds
UCI: Student Performance
UCI: Heart Disease Dataset
UCI: Bank Marketing Dataset
UCI: Forest Fires Dataset
UCI: Yeast Dataset

World Bank WDI
OECD PISA Scores
US Census ACS
US Bureau of Labor Statistics
US Bureau of Economic Analysis
UN COMTRADE
WHO Mortality Database
NHANES Survey Data
FRED Economic Data
US Energy Information Administration
Global Carbon Project
NOAA Climate Data
Berkeley Earth Temperature
Johns Hopkins COVID-19 Time Series
FAO Food Price Index
USDA Crop Production Data
OpenFlights Airport and Routes
\end{verbatim}
\end{minipage} \\
\bottomrule
\end{tabular}
\end{table}

\section{Use of LLMs}
In addition to conventional data collection and analysis, we made use of LLMs at several stages of our work. 
First, LLMs were applied during the writing process to assist with polishing and improving the clarity of the manuscript. 
Second, LLMs were also leveraged to support certain aspects of dataset construction, where they were used to generate and refine synthetic examples in a controlled manner. 
These uses were complementary to our primary methodology and were limited to auxiliary tasks such as language editing and expanding data diversity, without affecting the core experimental design or evaluation. 

\end{document}